\documentclass[journal]{IEEEtran}

\usepackage{amsmath}
\usepackage{amssymb}
\usepackage{amsfonts}
\usepackage{graphicx}
\usepackage{epsfig}
\usepackage{subfigure}
\usepackage{psfrag}
\usepackage{cite}
\usepackage{latexsym}
\usepackage{url}
\usepackage{color}
\usepackage{multirow}
\usepackage{mathtools}
\usepackage{bm}
\usepackage{booktabs}
\usepackage{algorithm}
\usepackage{algpseudocode}
\usepackage{bm}

\usepackage{verbatim}
\usepackage{indentfirst}
\usepackage{hyperref}
\usepackage{multirow}
\usepackage{bbm}

\graphicspath{{fig/}}

\PassOptionsToPackage{bookmarks={false}}{hyperref}

\IEEEoverridecommandlockouts
\newtheorem{definition}{\underline{Definition}}[section]

\newtheorem{lemma}{\underline{Lemma}}[section]

\newtheorem{proposition}{\underline{Proposition}}[section]

\newtheorem{remark}{\underline{Remark}}[section]
\newcommand{\mv}[1]{\mbox{\boldmath{$ #1 $}}}

\allowdisplaybreaks[4] 

\begin{document}
\title{Quantization-Aware Collaborative Inference for Large Embodied AI Models}
\author{
Zhonghao Lyu, Ming Xiao, \emph{Senior Member, IEEE,} Mikael Skoglund, \emph{Fellow, IEEE,}\\   M\'erouane Debbah, \emph{Fellow, IEEE,} and H. Vincent Poor, \emph{Life Fellow, IEEE}
\thanks{Z. Lyu, M. Xiao, and M. Skoglund are with the Department of Information Science and Engineering, KTH Royal Institute of Technology, Stockholm, Sweden (e-mail: lzhon@kth.se, mingx@kth.se, skoglund@kth.se).}
\thanks{M. Debbah is with  the Research Institute for Digital Future, Khalifa University, 127788 Abu Dhabi, UAE (email: merouane.debbah@ku.ac.ae) and also with CentraleSupelec, University Paris-Saclay, 91192 Gif-sur-Yvette, France.}
\thanks{H. V. Poor is with the Department of Electrical and Computer
Engineering, Princeton University, New Jersey 08544, USA (e-mail:
poor@princeton.edu).}
}

\maketitle

\begin{abstract}
Large artificial intelligence models (LAIMs) are increasingly regarded as a core intelligence engine for embodied AI applications. However, the massive parameter scale and computational demands of LAIMs pose significant challenges for resource-limited embodied agents. To address this issue, we investigate quantization-aware collaborative inference (co-inference) for embodied AI systems.
First, we develop a tractable approximation for quantization-induced inference distortion. Based on this approximation, we derive lower and upper bounds on the quantization rate-inference distortion function, characterizing its dependence on LAIM statistics, including the quantization bit-width.
Next, we formulate a joint quantization bit-width and computation frequency design problem under delay and energy constraints, aiming to minimize the distortion upper bound while ensuring tightness through the corresponding lower bound.
Extensive evaluations validate the proposed distortion approximation, the derived rate-distortion bounds, and the effectiveness of the proposed joint design. Particularly, simulations and real-world testbed experiments demonstrate the  effectiveness of the proposed joint design in balancing inference quality, latency, and energy consumption in edge embodied AI systems.
\end{abstract}

\section{Introduction}

\subsection{Research Background and Related Work}
As  mobile networks evolve toward artificial intelligence radio access networks (AI-RANs) that natively support edge intelligence \cite{kundu2025airantransformingranaidriven,10791415}, embodied AI has emerged as one of the most representative and demanding application paradigms. In embodied AI systems, intelligent agents form a tightly coupled closed-loop among perception, communication, understanding, and action, enabling them to accomplish complex tasks such as navigation, dialogue, and collaboration in dynamic environments \cite{gupta2021embodied}.
In recent years, large AI models (LAIMs) have demonstrated remarkable capabilities in visual understanding, language reasoning, and multimodal task processing. As a result, LAIMs are increasingly regarded as the core intelligence engine of embodied systems, providing autonomous platforms with more general and powerful perception-reasoning-action capabilities \cite{10648594}.

However, embodied intelligence applications typically impose stringent requirements on real-time responsiveness, energy efficiency, and reliable inference performance. In this context, the two conventional deployment paradigms for LAIM inference, namely, on-device (agent) deployment \cite{10906629,10.1145/3666025.3699355} and on-cloud deployment \cite{griggs2024melangecostefficientlarge,jiang2025thunderservehighperformancecostefficientllm,10.1145/3676641.3716025},  face fundamental limitations.
On one hand, pure on-device deployment is severely constrained by the limited computation capability, memory capacity, and energy budget of embodied agents, making it difficult to accommodate the massive parameter scale and computational intensity of LAIMs \cite{10906629,10.1145/3666025.3699355}.
On the other hand, on-cloud deployment offers abundant computational resources but is highly sensitive to communication network conditions. Moreover, massive concurrent inference requests may cause server-side congestion, making it difficult to guarantee timely response and stable interaction control \cite{10.1145/3676641.3716025}. Such centralized deployment also raises serious concerns regarding data privacy and security, especially for safety-critical embodied AI applications.

To address these challenges,  collaborative inference (co-inference) across device, edge, and cloud has emerged as a promising paradigm for supporting LAIM-enabled embodied intelligence services. By deploying LAIMs at the network edge, embodied agents can perform necessary front-end lightweight processing locally and transmit intermediate features to the edge or cloud for subsequent inference \cite{10591707,10.1145/3704413.3764429,chen2024adaptive,11140540,10818760,11301737}. Such a co-inference framework effectively alleviates on-device resource bottlenecks, enabling resource-limited agents to access LAIM services. Compared with pure on-cloud deployment, bringing LAIMs closer to data sources allows parts of the computation to be offloaded to agents, thereby better exploiting distributed computational resources while improving privacy preservation. Furthermore, for emerging multimodal inference tasks, transmitting raw sensory data to servers often incurs substantial communication overhead. In contrast, co-inference can  reduce communication overhead while improving robustness against channel impairments \cite{10854360}. This advantage becomes particularly pronounced in multi-agent
and agentic AI scenarios \cite{jiang2025largeaimodelsagentic}, where structured representations facilitate feature-level information fusion across agents at the edge or cloud, leading to more comprehensive situational awareness and enhanced task performance.

Existing studies on co-inference can be broadly categorized into two research directions: task-oriented feature encoding and joint task offloading and resource allocation.
Task-oriented feature encoding, also referred to as semantic or goal-oriented communications, focuses on transforming raw data into semantic representations prior to transmission,  to reduce communication overhead while supporting downstream tasks. This paradigm has been applied to multimodal data reconstruction \cite{10854360,11038757,soret2024semanticgoalorientededgecomputing}, intelligent task execution \cite{10644029,10016643}, and multi-task inference services \cite{10388062}.
In parallel, joint task offloading and resource allocation studies aim to partition and dynamically assign inference workloads across distributed processing entities according to heterogeneous system resources \cite{ren2023survey}. Representative works have investigated adaptive co-inference frameworks that respectively optimize data representation and workload partitioning to balance inference performance and resource efficiency under constrained edge environments \cite{li2025taskorientedcomputationoffloadingedge,9296560}. Moving toward multi-device settings, \cite{9837474,10829586,11121577} have further explored information-bottleneck based feature transmission, over-the-air computation, and sensing-communication-computation co-design for multi-view, privacy-preserving, and perceptive co-inference scenarios.

However, the above studies mainly focused on regular-scale AI models. For LAIMs, the dramatically increased model scale and computational demand necessitate large-scale system design strategies. For LAIM-oriented co-inference, \cite{10591707} has proposed a cloud-edge active co-inference framework, yet neglected the computation capability of edge devices. Subsequently, \cite{10.1145/3704413.3764429,chen2024adaptive} have investigated multimodal, multi-task, and multi-dialogue LAIM task-offloading and dynamic LAIM splitting schemes 
using deep reinforcement learning (DRL). Hybrid large-small language model architectures have also been explored to enable selective uplink transmission and cloud invocation for LAIM inference \cite{11140540}. More recently, LAIM co-inference has also been extended to multi-device multi-server settings with joint device selection and model partitioning \cite{10818760}.

Despite these advances, for LAIM co-inference in embodied AI, the on-agent component still remains fundamentally constrained by limited resources, rendering model compression indispensable for practical deployment \cite{sreenivas2024llmpruningdistillationpractice}. To this end, several compression techniques have been studied, including pruning, quantization, and knowledge distillation.
Model pruning removes redundant parameters, neurons, or attention heads based on magnitude, sensitivity, or structural criteria \cite{11301737}. Knowledge distillation transfers knowledge from a large teacher model to a compact student model by aligning output distributions or intermediate representations \cite{gu2024minillm}.
In contrast, model quantization reduces numerical precision by representing model parameters and activations with low bit-width formats, substantially decreasing storage and computation complexity with relatively low engineering overhead, while maintaining compatibility with standard edge hardware. Widely adopted quantization schemes include uniform quantization \cite{10.5555/3045390.3045690} and nonuniform quantization \cite{zhou2017incremental}, where quantization levels are uniformly or nonuniformly distributed, respectively. As a result, quantization has emerged as a particularly attractive solution for LAIM co-inference on resource-limited embodied agents.


\subsection{Motivations and Contributions}
Despite the extensive studies on model quantization algorithms, quantization inevitably introduces parameter precision degradation that leads to inference distortion, and a fundamental gap remains in understanding how such distortion scales with the quantization bit-width in a tractable  manner. Existing works largely lack a unified theoretical characterization of the quantization-distortion trade-off that can guide system-level optimization.
More importantly, from a system perspective, quantization affects not only inference accuracy, but also the computational workload. Meanwhile, computation frequency control at both device and server processors constitutes another critical design degree of freedom (DoF) that determines inference delay and energy consumption. As a result, quantization bit-width and computation frequency become tightly coupled under practical quality-of-service (QoS) constraints. For instance, reducing the bit-width can lower computational load at the cost of increased inference distortion, whereas increasing the computation frequency may shorten inference latency but amplify energy consumption. This coupling gives rise to non-trivial inference quality-energy-delay trade-offs, which are particularly important in heterogeneous and resource-constrained embodied AI systems. Balancing these trade-offs calls for quantization-aware distortion characterization and joint model quantization and computation design. However, such issues remain largely unexplored in existing works.

Motivated by the above gaps, this paper adopts a unified perspective that bridges theoretical analysis and system-level design, and develops a joint quantization and computation design framework for LAIM-enabled co-inference. Specifically, we first construct a  quantization-induced distortion metric and perform a rate-distortion analysis to characterize the fundamental relationship between quantization bit-width and inference distortion. Then, we jointly optimize the on-agent quantization bit-width and the computation frequencies under QoS constraints, providing practical design insights for model compression and computation resource management in embodied AI systems. The main contributions are summarized as follows.
\begin{itemize}
	\item We develop an efficient approximation for quantization-induced inference distortion in LAIMs by explicitly linking parameter-level perturbations to output-level distortion. Building on this approximation, we derive upper and lower bounds on inference distortion from a rate-distortion theoretic perspective, which fundamentally  characterize how inference quality depends on key LAIM statistics, such as the quantization bit-width. 
	\item Leveraging the above analysis, we formulate a joint quantization bit-width and computation frequency optimization problem under delay and energy constraints. The objective is to minimize the distortion upper bound while maintaining a tight approximation gap, achieved by jointly maximizing the corresponding lower bound.
	The problem is non-convex due to the coupling among decision variables, for which we propose an efficient solution.
    \item Finally, we conduct extensive evaluations to validate both the  theoretical analysis and system-level design. The results verify  the quantization-induced distortion approximation and the derived rate-distortion bounds. In particular, experiments on multiple LAIMs, datasets, and quantization schemes demonstrate consistent performance gains in balancing inference quality, latency, and energy consumption. Real-world testbed results further confirm that, even with only coarse-grained frequency configurations, jointly optimizing computation and quantization remains crucial for improving LAIM co-inference performance in practical edge embodied AI systems.
\end{itemize}

The remainder of this paper is organized as follows. Section II introduces the system model. Section III approximates the LAIM output distortion. Section IV analyzes the rate-distortion relationship for quantization. Section V jointly designs quantization and computation. Section VI reports the evaluation results. Section VII concludes the paper.

\begin{figure}[h]
	\centering
	 \epsfxsize=1\linewidth
		\includegraphics[width=8.5cm]{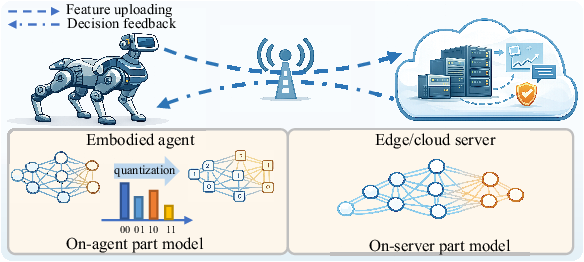}
	\caption{\label{fig:system_model}The considered LAIM-enabled embodied AI system.}
\end{figure}

\section{System Model}

We consider a quantization-aware LAIM co-inference embodied AI system consisting of one agent and one server, as illustrated in Fig.~\ref{fig:system_model}. Specifically, the inference process is divided into two sequential stages. In the first stage, the agent performs partial inference on the input data to extract intermediate embeddings, which are then transmitted to the server. In the second stage, the  server completes the remaining inference process based on the received embeddings and returns the final inference results to the agent. 

\subsection{On-agent Inference and Embedding Transmission}

We first describe the on-agent inference process. Let $\mv{w} \in \mathbb{R}^{q}$ denote the parameter vector of the on-agent model, where each parameter is originally stored using $b \in \mathbb{Z}_{+}$ bits. To enable efficient inference under resource constraints, the on-agent model is quantized using a reduced bit-width $\hat b \in \mathcal{B} \triangleq \{1, \ldots, B_{\max}\} \subset \mathbb{Z}_{+}$,  where $B_{\max}$ denotes the maximum allowable bit-width. This quantization results in a lightweight on-agent model $\hat{\mv{w}} \in \mathbb{R}^{q}$.

Let $\mv{x}$ denote the input data. The on-agent inference output, i.e., the intermediate embedding, is given by
\begin{align}
    \mv{o} = f(\mv{x}, \hat{\mv{w}}),
\end{align}
where $f(\cdot)$ represents the computation function of the quantized on-agent model $\hat{\mv{w}}$. The embedding $\mv{o}$ is then transmitted to the edge server  for further processing.

\subsection{On-server Inference and Result Feedback}

Upon receiving the embedding $\mv{o}$, the server executes the remaining inference process. Let $\mv{v} \in \mathbb{R}^{s}$ denote the parameter vector of the on-server model. The final inference output is obtained as
\begin{align}
    \tilde{\mv{o}} = \tilde{f}(\mv{o}, \mv{v}),
\end{align}
where $\tilde{f}(\cdot)$ denotes the computation function of the on-server model $\mv{v}$. The server subsequently feeds back the final inference result $\tilde{\mv{o}}$ to the agent, thereby completing the collaborative inference process.

\subsection{LAIM Parameter Distribution Modeling}

To facilitate tractable performance analysis, we model the statistical distribution of LAIM parameters. Specifically, we preserve the sign bits of model parameters and quantize only their magnitudes. Denote $\mv{\theta}=|\mv{w}|$ as the magnitude of the parameter vector $\mv{w}$. We assume that the magnitude of each parameter element $\theta \in \mv{\theta}$ follows an exponential distribution with probability density function (PDF)
\begin{align}\label{PDF}
    P_{\mathit{\Theta}}(\theta) = \lambda e^{-\lambda \theta}, \quad \theta \ge 0,
\end{align}
where $\lambda > 0$ is the distribution parameter.

\begin{figure*}[h]
    \centering 
	\subfigure[ResNet-152]{\includegraphics[width=.3\textwidth]{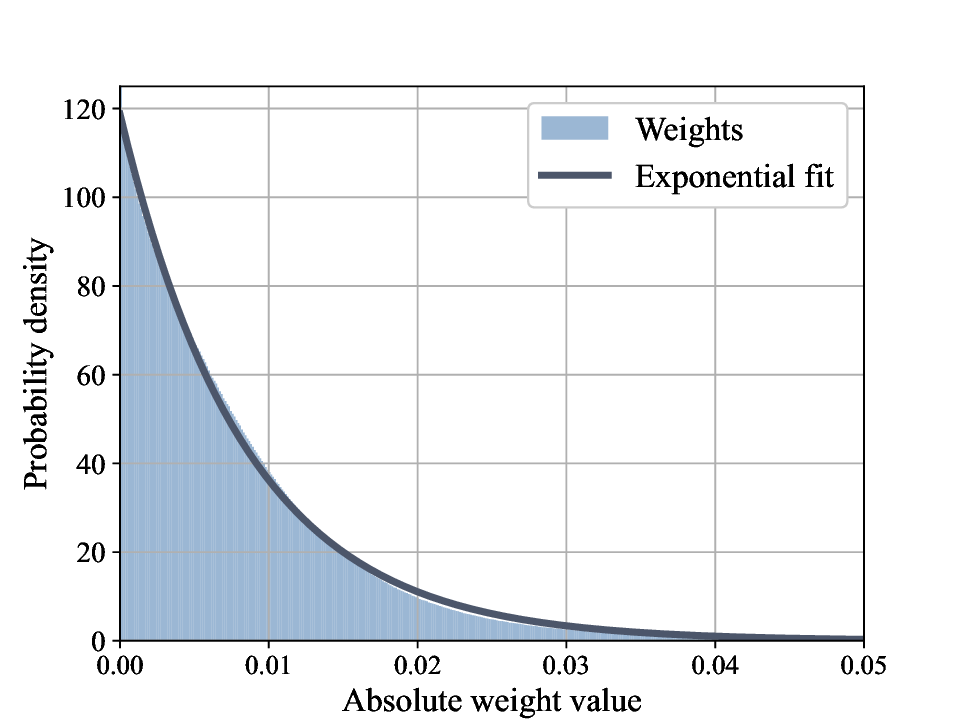}}
    \subfigure[VideoMAE]{\includegraphics[width=.3\textwidth]{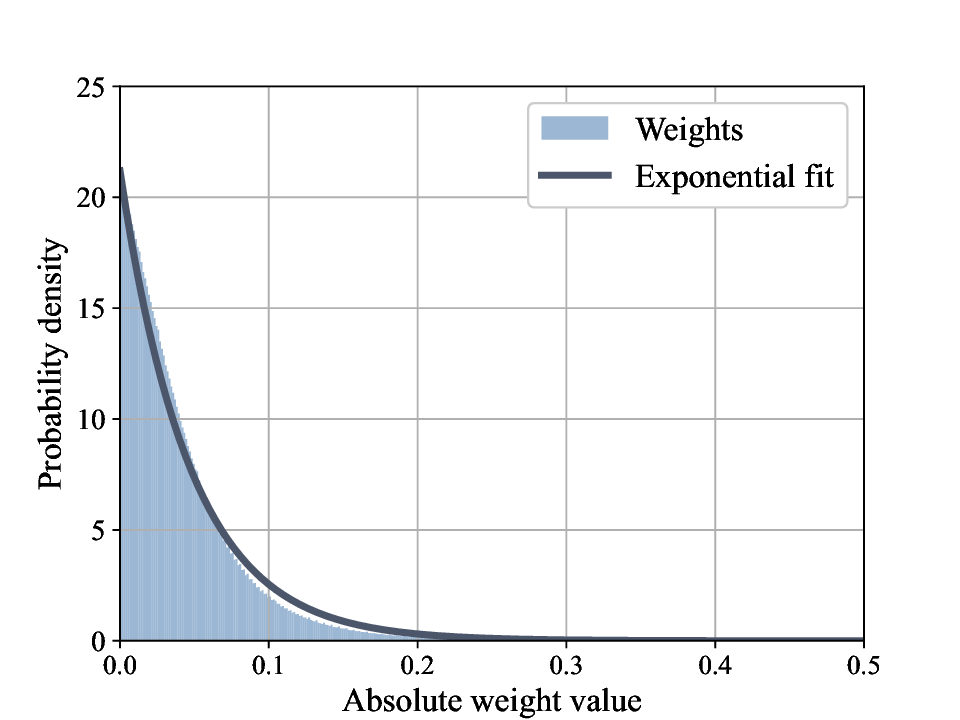}}
    \subfigure[BERT]{\includegraphics[width=.3\textwidth]{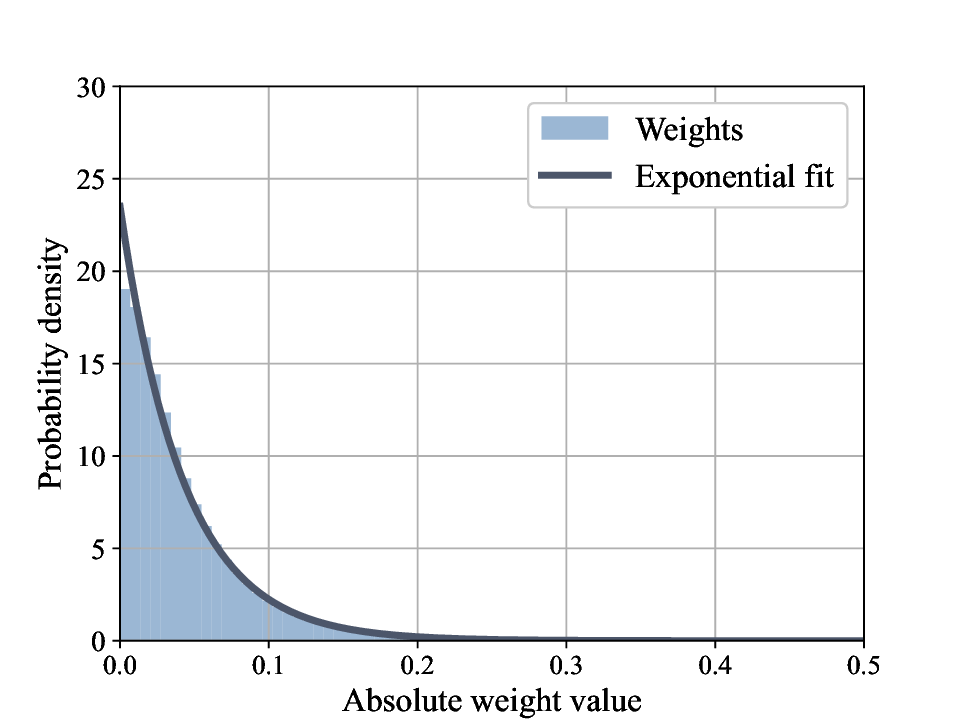}}
	\\
	\subfigure[BLIP-2]{\includegraphics[width=.3\textwidth]{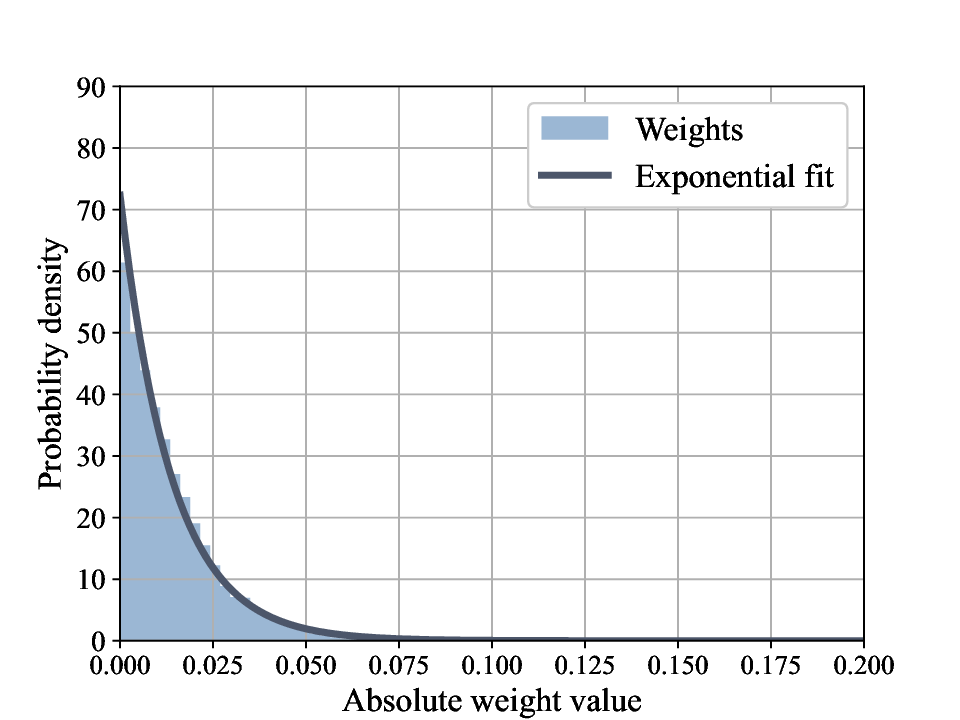}}
	\subfigure[GIT]{\includegraphics[width=.3\textwidth]{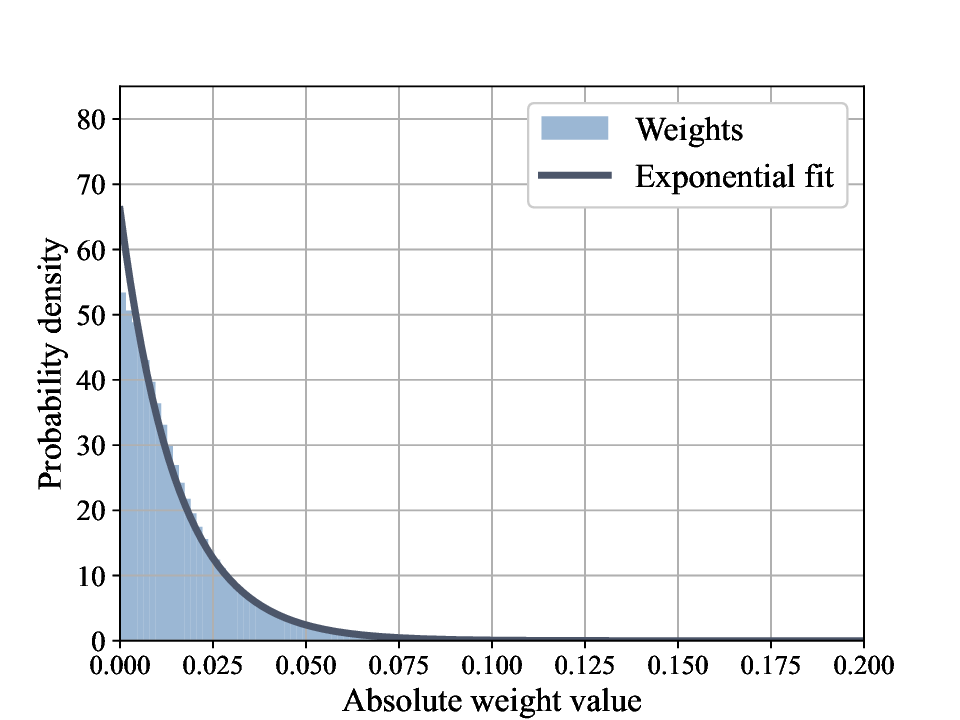}}
	\subfigure[GPT-3]{\includegraphics[width=.3\textwidth]{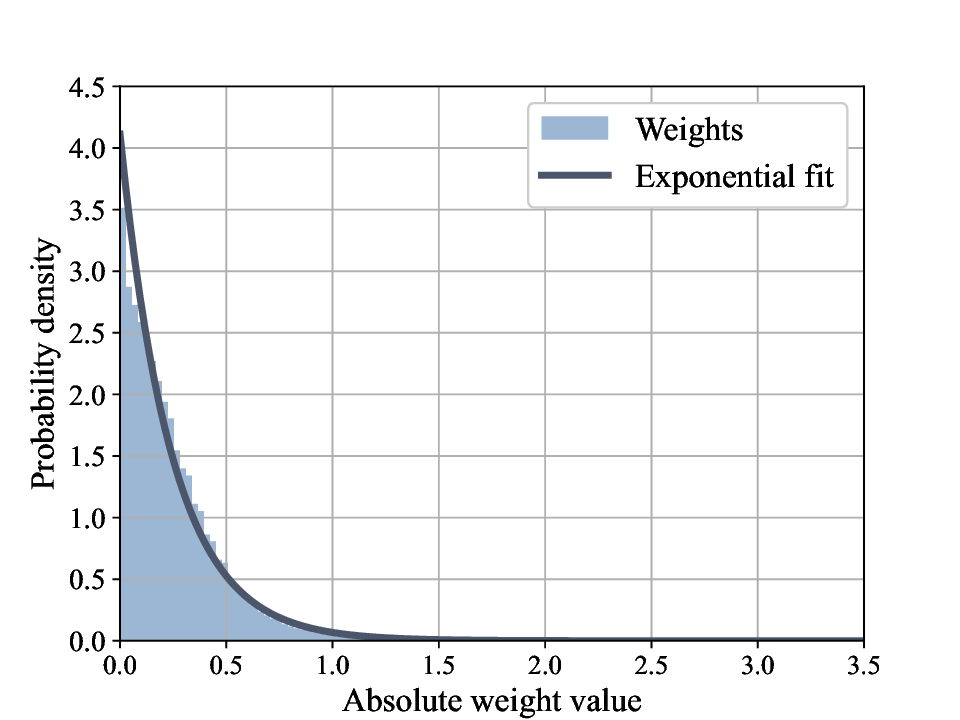}}
    \caption{Distribution of the parameter magnitudes of various pre-trained models.}
    \label{fig:density_weight}
\end{figure*}

This modeling assumption is empirically supported as shown in Fig. \ref{fig:density_weight}. 
Specifically, we consider the following models, ResNet-152 \cite{7780459}, VideoMAE \cite{10.5555/3600270.3601002}, BERT \cite{devlin-etal-2019-bert}, BLIP-2 \cite{10.5555/3618408.3619222}, GIT \cite{wang2022git}, and GPT-3 \cite{10.5555/3495724.3495883}, which together represent a diverse set of modalities and model families. 
For each model, we collect the corresponding weights from the pre-trained checkpoints and compute their magnitudes to form the empirical histogram for distribution fitting, where the exponential distribution is used as the candidate model. The empirical distributions of the parameter magnitudes closely match the exponential distribution in Fig. \ref{fig:density_weight}, which shows a sharp peak of weight values around zero. Accordingly, we assume that the parameters of a pre-trained LAIM are independently drawn from this distribution.

\subsection{Inference Delay and Energy Consumption}

We analyze the end-to-end inference latency and energy consumption of the co-inference system. Since LAIM inference is typically computation-dominated, we focus on computation-related delay and energy consumption.

\subsubsection{Delay Analysis}

Let $N_{\rm FLOP}$ denote the number of floating-point operations (FLOPs) required for full-precision on-agent inference. After quantization, we assume that the effective computational workload scales linearly with the quantization bit-width, resulting in a workload of $\hat b N_{\rm FLOP}/b$. Let $f$ and $c$ denote the clock frequency and the number of FLOPs per CPU cycle of the device processor, respectively. The on-agent inference delay is then given by
\begin{align}
    t(\hat b, f) = \frac{\hat b N_{\rm FLOP}}{b f c}.
\end{align}

Similarly, let $\tilde{N}_{\rm FLOP}$ denote the total computational workload required for the on-server inference stage. Let $\tilde{f}$ and $\tilde{c}$ denote the clock frequency and FLOPs per cycle of the server processor, respectively. The on-server inference delay is
\begin{align}
    \tilde{t}(\tilde{f}) = \frac{\tilde{N}_{\rm FLOP}}{\tilde{f} \tilde{c}}.
\end{align}

\subsubsection{Energy Consumption Analysis}

Let $\eta$ and $\psi$ denote the power usage effectiveness (PUE) and a chip-dependent power coefficient of the agent, respectively. The energy consumption of on-agent inference is modeled as
\begin{align}
    e(\hat b, f) = \eta \frac{\hat b N_{\rm FLOP}}{b c} \psi f^2.
\end{align}

Similarly, let $\tilde{\eta}$ and $\tilde{\psi}$ denote the PUE and power coefficient of the server. The energy consumption of on-server inference is given by
\begin{align}
    \tilde{e}(\tilde{f}) = \tilde{\eta} \frac{\tilde{N}_{\rm FLOP}}{\tilde{c}} \tilde{\psi} \tilde{f}^2.
\end{align}

Finally, the total inference computational delay and energy consumption are respectively expressed as
\begin{align}
    T(\hat b, f, \tilde{f}) &= t(\hat b, f) + \tilde{t}(\tilde{f}), \\
    E(\hat b, f, \tilde{f}) &= e(\hat b, f) + \tilde{e}(\tilde{f}).
\end{align}

\section{Model Output Distortion Approximation}

To design  LAIM co-inference in embodied AI systems, it is essential to characterize the inference output distortion introduced by model quantization. Such a characterization is challenging due to the highly nonlinear structure and depth of LAIMs. Directly analyzing the exact relationship between parameter quantization and inference output distortion is generally intractable. To address this challenge, we start by studying the output distortion of fully connected (FC) deep neural networks (DNNs) under model quantization, which provides explicit insights into how parameter perturbations propagate through layered architectures and affect inference outputs. Then, we extend the discussions to general AI models, including transformer-based LAIMs, to justify the use of a tractable surrogate distortion measurement (i.e., parameter distortion) for subsequent quantization-enabled co-inference analysis and optimization.

First, we approximate the output distortion of an $L$-layer FC DNN under model quantization. Denote $\mv{W}=\{\mv{W}^{(1)},\ldots,\mv{W}^{(L)}\}$ as the collection of $L$ layers weight matrices, and $\sigma(\cdot)$ as the activation function. Then, the FC DNN model is expressed as
\begin{align}
	f(\mv{x},\mv{W}) = \mv{W}^{(L)} \sigma\!\left(\mv{W}^{(L-1)} \sigma\big(\cdots \mv{W}^{(1)} \mv{x}\big)\right).
\end{align}

Then, to facilitate tractable distortion analysis, we introduce the following assumptions on the input data, activation function, and quantization errors.

\textbf{Assumption 1} (Normalized input).  
The input data are normalized such that $\|\mv{x}\|_1^2 \le 1$.

\textbf{Assumption 2 }  (Activation smoothness).
The activation function $\sigma(\cdot)$ is 1-Lipschitz, i.e.,
\begin{align}
	\|\sigma(\mv{x}) - \sigma(\mv{y})\|_1 \le \|\mv{x} - \mv{y}\|_1,
\end{align}
and satisfies $\sigma(0)=0$. This assumption holds for widely used activation functions such as ReLU, LeakyReLU, and tanh.

\textbf{Assumption 3 }  (Bounded quantization error).
The quantization error of the $l$-th layer is bounded as
\begin{align}
	\|\mv{W}^{(l)} - \hat{\mv{W}}^{(l)}\|_1 \le \tau^{(l)},
\end{align}
where $\tau^{(l)}$ depends on the quantization resolution/level.

Under these assumptions, we establish the following upper bound on the output distortion induced by model quantization.

\begin{proposition}\label{proposition1}
Under Assumptions 1-3, an upper bound on the inference output distortion of an FC DNN induced by model quantization is given by
\begin{align}
	\|f(\mv{x},\mv{W}) - f(\mv{x},\hat{\mv{W}})\|_1 
	\le \sum_{l=1}^{L} A^{(l)} \|\mv{W}^{(l)} - \hat{\mv{W}}^{(l)}\|_1,
\end{align}
where
\begin{align}
	A^{(l)} = \prod_{j=1}^{l-1} \|\mv{W}^{(j)}\|_1 
	\prod_{k=l+1}^{L} \big(\|\mv{W}^{(k)}\|_1 + \tau^{(k)}\big).
\end{align}
\begin{IEEEproof}
	See Appendix~A.
\end{IEEEproof}
\end{proposition}

\begin{remark}[Distortion metric]
Proposition~\ref{proposition1} shows that the inference output distortion induced by quantization is upper bounded by a weighted sum of layer-wise parameter distortions. This result establishes a  theoretical link between model-level quantization perturbations and task-level inference distortion. Importantly, the multiplicative coefficients $A^{(l)}$ depend only on the unquantized model parameters and the quantization errors, and are independent of the quantized model parameters.

This observation motivates the use of a norm-based parameter distortion as a surrogate metric for inference output distortion. In particular, we omit $A^{(l)}$ and adopt the following distortion function for the subsequent rate-distortion analysis,
\begin{align}\label{distortion}
	d(\mv{W}, \hat{\mv{W}}) 
	\!\!=\!\! \|\mv{W} - \hat{\mv{W}}\|_1 
	\!\!=\!\! \sum_i |w_i - \hat{w}_i|
	\!\!=\!\! \sum_i |\theta_i - \hat{\theta}_i|,
\end{align}
\end{remark}
where $w_i$, $\hat{w}_i$, $\theta_i$, $\hat{\theta}_i$ are elements of the weight matrix $\mv W$, the quantized weight matrix $\hat{\mv{W}}$, the magnitude  of the weight matrix $\bm{\mathit{\Theta}}=|\mv W|$, and the  magnitude  of the quantized weight matrix $\hat{\bm{\mathit{ \Theta}}}=|\hat{\mv{W}}|$, respectively.

\begin{remark}[Extension to General AI Models]
The distortion measure in \eqref{distortion} is also applicable to general AI models, including transformer-based LAIMs, where explicit expressions of inference outputs (i.e., $f(\mv{x},\mv{W})$ and $f(\mv{x},\hat{\mv{W}})$) and the corresponding quantization-induced distortion are analytically intractable due to the model's high structural complexity. To tackle this issue, we approximate the output after quantization $f(\mv{x},\hat{\mv{W}})$ with the first-order Taylor expansion as
\begin{align}\label{taylor}
	f(\mv{x},\hat{\mv{W}}) 
	\approx f(\mv{x},\mv{W}) 
	+ \nabla_{\mv W} f(\mv{x},\mv{W}) 
	: (\hat{\mv W} - \mv{W}),
\end{align}
where $:$ denotes tensor contraction. Furthermore, we assume the gradient is upper bounded as $\|\nabla_{\mv W} f(\mv{x},\mv{W})\|_{1} \le H$ by a positive constant $H>0$, then the quantization-induced output distortion is approximately upper bounded as
\begin{align}\label{distortion of non-linear}
	\|f(\mv{x},\hat{\mv{W}}) - f(\mv{x},\mv{W}) \|_1
	\le H \|\mv{W}-\hat{\mv{W}}\|_1.
\end{align}
This result indicates that, even for nonlinear and large-scale models, the inference output distortion caused by quantization is related to a norm of the parameter perturbation. Consequently, the distortion function in \eqref{distortion} serves as a theoretically justified surrogate for output distortion metric in subsequent rate-distortion analysis, enabling tractable characterization of the relationship between quantization bit-width and LAIM co-inference performance.
\end{remark}

\section{Rate-distortion Analysis for Quantization}

This section develops an information-theoretic characterization of the fundamental trade-off between model quantization bit-width and inference distortion for LAIM co-inference in embodied AI. As discussed in Section~III, directly analyzing the exact inference output distortion of LAIMs is  intractable. Motivated by Proposition~\ref{proposition1}, we instead study the rate-distortion behavior of the parameter distortion under the assumed exponential weight distribution and distortion metric  in \eqref{distortion}. 
Furthermore, deriving the exact rate-distortion function under our considered setup is generally non-trivial. Hence, we derive a tractable lower bound and an upper bound on the rate-distortion function, which together provide a computable approximation for the rate-distortion relationship. The analyzed bounds are exploited in Section~V for quantization-aware system design.

\subsection{Preliminaries of Rate-distortion Analysis}

We briefly recall the definition of the rate-distortion function and specialize it to model quantization.

\begin{definition}[Rate-distortion function \cite{cover1999elements,733495}]
Consider a memoryless real-valued source sequence $\mathit{\Theta}=\{\theta_1,\ldots,\theta_n\}$, where each sample $\theta_i \in \mathbb{R}$ is drawn independently and identically according to a PDF $P_\mathit{\Theta}$. 
An $(n,R)$ source code consists of an encoder $e_n:\mathbb{R}^n \to \{1,\ldots,2^{nR}\}$ and a decoder $g_n:\{1,\ldots,2^{nR}\}\to\mathbb{R}^n$.
Let $\hat {\mathit{\Theta}} \triangleq g_n(e_n(\mathit{\Theta}))$ denote the reconstruction.
For a distortion measure $d(\mathit{\Theta},\hat{ \mathit{\Theta}})\triangleq \frac{1}{n}\sum_{i=1}^n d(\theta_i,\hat \theta_i)$, the rate-distortion function is defined as
\begin{align}
R(D) = \inf_{P_{\hat {\mathit{\Theta}}|\mathit{\Theta}}:\,\mathbb{E}[d(\mathit{\Theta},\hat {\mathit{\Theta}})] \le D} I(\mathit{\Theta};\hat {\mathit{\Theta}}),
\label{eq:RD-def-scalar}
\end{align}
where the infimum is over all conditional distributions $P_{\hat {\mathit{\Theta}}|\mathit{\Theta}}$.
Accordingly, the distortion-rate function $D(R)$ is defined as the infimum of all $D$ such that \eqref{eq:RD-def-scalar} does not exceed a given rate $R$.

In the context of model quantization on parameter magnitude $\theta$, consistent with the surrogate distortion in \eqref{distortion}, we adopt the  distortion measurement as $d(\theta,\hat \theta) = |\theta-\hat \theta|$.
Then $R(D)$ characterizes the minimum number of bits (bit-width) per model parameter required to achieve an average quantization distortion no larger than $D$.
\end{definition}

\subsection{A Lower Bound on the Rate-distortion function for Quantization}

We derive a Shannon-type lower bound on the rate-distortion function under the considered exponential distribution and distortion function in \eqref{distortion}. First, we introduce the following lemmas.

\begin{lemma}\label{lemma1}
Denote the quantized magnitude of model parameter as ${ \mathit{ \hat \Theta}} \triangleq \mathit{\Theta}+ Z$, where $Z$ is the quantization noise. Assume $Z$ is a real-valued random variable with PDF $P_Z$. Then the rate-distortion function is lower bounded by
\begin{align}\label{lemma-RD}
R(D) \ge h(\mathit{\Theta}) - \Phi(D),
\end{align}
where
\begin{align}
\Phi(D) \triangleq \sup_{P_Z:\,\mathbb{E}[|Z|]\le D} h(Z),
\end{align}
and $h(\mathit{\Theta})$ denotes the differential entropy of $\mathit{\Theta}$, i.e.,
\begin{align}
h(\mathit{\Theta}) = -\int P_{\mathit{\Theta}}(\theta)\log_2(P_{\mathit{\Theta}}(\theta))d\theta=\log_2 ({\rm e}/ \lambda).
\end{align}
\end{lemma}

\begin{lemma}\label{lemma2}
Among all $Z$ with finite first absolute moment and satisfying $\mathbb{E}[|Z|]\le D$, the differential entropy is maximized by a zero-mean Laplacian distribution $Z_D$ with PDF $P_{Z_D}(z) = \frac{1}{2D}{\rm e}^{-|z|/D}$, for which
\begin{align}
h(Z_D) = \log_2(2{\rm e}D).
\end{align}
\begin{IEEEproof}
See Appendix B.
\end{IEEEproof}
\end{lemma}

\begin{proposition}\label{prop:slb-exponential}
For the exponentially distributed LAIM parameters with PDF as \eqref{PDF}, under the distortion metric $d(\theta,\hat \theta)=|\theta-\hat \theta|$, the rate-distortion function is lower bounded by
\begin{align}
R(D) \ge  -\log_2(2\lambda D) \triangleq R^{\rm L}(D).
\label{eq:SLB-exp-RD}
\end{align}
Equivalently, the distortion-rate function is lower bounded by
\begin{align}
D(R) \ge  \frac{1}{\lambda 2^{R+1}} \triangleq D^{\rm L}(R).
\label{eq:SLB-exp-DR}
\end{align}
\end{proposition}

\begin{remark}
Proposition~\ref{prop:slb-exponential} provides a clear relationship between quantization bit-width and distortion. 
Specifically, \eqref{eq:SLB-exp-DR} shows increasing bit-width generally yields distortion reduction, however, with increased computation workload as well as delay and energy consumption. Thus, it is crucial to  design quantization bit-width with computation resources to balance the trade-offs among inference quality, latency, and energy consumption.
Second, $\lambda$ (or equivalently the entropy level of the weight distribution) serves as an indicator of how quantization-sensitive a given model is. Specifically, the parameter $\lambda$ captures the concentration of model weights near zero. A larger $\lambda$ implies a sharper peak at zero and hence a smaller bit-width is needed for the same distortion, whereas smaller $\lambda$ corresponds to a heavier tail and higher distortion sensitivity. This motivates incorporating model statistics into quantization-aware co-inference design.
\end{remark}

\subsection{An Upper Bound on the Rate-distortion Function for Quantization}

We further provide an upper bound on the rate-distortion function as follows.

\begin{proposition}\label{prop:sub-exponential}
For the exponentially distributed LAIM parameters with PDF as \eqref{PDF}, under the distortion metric $d(\theta,\hat \theta)=|\theta-\hat \theta|$, the rate-distortion function is upper bounded by
\begin{align}
R(D) \le  \log_2\!\left(\frac{1}{\lambda D}+\frac{\lambda D}{\lambda D+1}\right) \triangleq R^{\mathrm{U}}(D).
\label{eq:UB-exp-RD}
\end{align}
Equivalently, the distortion-rate function is upper bounded as
\begin{align}
D(R) \le \frac{1}{2 \lambda}\!\left(\sqrt{1+\frac{4}{2^R-1}}-1\right) \triangleq D^{\rm U}(R).
\label{eq:UB-exp-DR}
\end{align}
\end{proposition}

\begin{IEEEproof}
First, we construct a valid test channel and upper bound $R(D)$ by evaluating $I(\mathit{\Theta};\mathit{\hat \Theta})$, i.e., 
\begin{align}
R(D) &\le I(\mathit{\Theta};\mathit{\hat \Theta}) 
= I(\mathit{\Theta};\mathit{\Theta}+Z) \nonumber \\
&= h(\mathit{\Theta}+Z) - h(\mathit{\Theta}+Z|\mathit{\Theta}) \nonumber \\
&= h(\mathit{\Theta}+Z) - h(Z). \label{eq:app-ub1}
\end{align}
By Lemma~\ref{lemma2}, the Laplacian distribution could tighten the upper bound in \eqref{eq:app-ub1}. Thus, we choose $Z$ as a zero-mean Laplacian random variable independent of $\mathit{\Theta}$ and satisfying $\mathbb{E}[|Z|]=D$. Then we have
\begin{align}\label{RD_upper}
R(D) 
&\le \log_2\!\big(2{\rm e}\,\mathbb{E}[|\mathit{\Theta}+Z|]\big) - \log_2(2{\rm e}D).
\end{align}
Finally, it remains to compute $\mathbb{E}[|\mathit{\Theta}+Z|]$ as
\begin{align}\label{E_XW}
&\mathbb{E}[|\mathit{\Theta}+Z|] \nonumber \\
&=\int_{0}^{\infty} \int_{-\infty}^{\infty} |\theta+z| f_{\mathit{\Theta}}(\theta) f_Z(z) dz d\theta \nonumber \\
&= \int_{0}^{\infty} \bigg( \int_{-\infty}^{-\theta} \frac{-(\theta+z)}{2D}e^{z/D} dz + \int_{-\theta}^{0} \frac{(\theta+z)}{2D}e^{z/D} dz \nonumber \\
& \qquad + \int_{0}^{\infty} \frac{(\theta+z)}{2D}e^{-z/D} dz \bigg) \lambda e^{-\lambda \theta}  d\theta \nonumber \\
& =\int_{0}^{\infty} (\theta+De^{-\theta/D}) \lambda e^{-\lambda \theta}  d\theta \nonumber \\
&= \frac{1}{\lambda} + D\cdot \frac{\lambda D}{\lambda D+1}.
\end{align}
Substituting \eqref{E_XW} into \eqref{RD_upper} gives \eqref{eq:UB-exp-RD}, which completes the proof.
\end{IEEEproof}

\begin{remark}
The bounds \eqref{eq:SLB-exp-DR} and \eqref{eq:UB-exp-DR} provide a computable interval $D^{\rm L}(R) \le D(R) \le D^{\rm U}(R)$
that captures the fundamental trade-off between quantization bit-width and distortion. This interval is particularly useful for LAIM co-inference system design. It yields a conservative distortion estimate (via $D^{\rm U}(R)$) and an optimistic distortion floor (via $D^{\rm L}(R)$), thereby quantifying the quantization-induced inference distortion. In Section V, we will leverage these two bounds for quantization-aware co-inference system design under QoS constraints.
\end{remark}

\section{Joint Quantization and Computation Design for LAIM Co-Inference}
This section jointly designs the quantization bit-width and the computation frequencies for LAIM co-inference in embodied AI systems. 

\subsection{Problem Formulation}
We minimize the quantization-induced performance distortion using the derived upper bound, while guaranteeing approximation tightness through the corresponding lower bound, subject to the delay and energy constraints of LAIM co-inference. The decision variables include the on-agent quantization bit-width $\hat b$  and the computation frequencies of the device and server, i.e., $f$ and $\tilde f$. The resulting optimization problem is formulated as follows:
\begin{subequations}\label{P1}
\begin{align}
\text{(P1)}:\ \min_{\hat b,f,\tilde f}\quad 
& D^{\rm U}(\hat b-1) -  D^{\rm L}(\hat b-1) \nonumber \\
\text{s.t.}\quad 
& T(\hat b,f,\tilde f) \le T_0, \label{P1-b}\\
& E(\hat b,f,\tilde f) \le E_0, \label{P1-c}\\
& \hat b \in \mathcal{B}, \label{P1-d}\\
& 0 \le f \le f^{\max}, \label{P1-f}\\
& 0 \le \tilde f \le \tilde f^{\max}, \label{P1-g}
\end{align}
\end{subequations}
where $T_0$ and $E_0$ denote the maximum allowable computation delay and energy consumption, respectively. Constraints \eqref{P1-f} and \eqref{P1-g} specify feasible computation frequency ranges, and \eqref{P1-d} enforces the achievable bit-width set. 
However, in problem (P1), the constraints in \eqref{P1-b} and \eqref{P1-c} are non-convex, due to the close coupling of variables $\hat b$, $f$, and $\tilde f$. Furthermore, the integer variable for quantization bit-width $\hat b$ makes problem (P1) a mixed-integer non-linear problem
(MINLP). Therefore, problem (P1) is highly non-convex and non-trivial to be optimally solved in general.

\begin{algorithm}[h]
\caption{Algorithm for Solving Problem (P1)}
\label{alg:sca_p1}
\begin{algorithmic}[1]
\State Relax $\hat b$ to $\tilde b$.
\State Initialize a feasible local point $(\tilde b^{(0)},\tilde b'^{(0)})$.
\State Set $k=1$.
\Repeat
\State Solve problem (P4.$k$) under $\tilde b^{(k-1)}$ and $\tilde b'^{(k-1)}$ to obtain the solution $\tilde b^{(k)*},\tilde b'^{(k)*},f^{(k)*}$, and $\tilde f^{(k)*}$.
\State Update $\tilde b^{(k)},\tilde b'^{(k)}\leftarrow \tilde b^{(k)*},\tilde b'^{(k)*}$.
\State Update $k\leftarrow k+1$.
\Until the decrease of the objective value is below a threshold $\delta$.
\State Obtain $(\tilde b^\star,f^\star,\tilde f^\star)$ from the final iteration; Round $\tilde b^\star$ to the nearest value $\hat b^\star$ in $\mathcal{B}$.
\State \textbf{Output} $(\hat b^\star,f^\star,\tilde f^\star)$.
\end{algorithmic}
\end{algorithm}

\subsection{Proposed Solution to Problem (P1)}

To handle the discrete variable $\hat b$, we relax it to a continuous variable $\tilde b \in (1,B_{\max}]$. Then problem (P1) is relaxed to
\begin{subequations}\label{P2}
\begin{align}
\text{(P2)}:\ \min_{\tilde b,f,\tilde f}\quad 
& \frac{1}{2\lambda}\!\left(\sqrt{1+\frac{4}{2^{\tilde b-1}-1}}-1\right)
 - \frac{1}{\lambda 2^{\tilde b}} \nonumber \\
\text{s.t.}\quad 
& \frac{\tilde b N_{\rm FLOP}}{b f c}+ \frac{\tilde N_{\rm FLOP}}{\tilde f \tilde c} \le T_0, \label{P2-a}\\
& \eta \frac{\tilde b N_{\rm FLOP}}{b c}\psi f^2 
+ \tilde\eta \frac{\tilde N_{\rm FLOP}}{\tilde c}\tilde\psi \tilde f^2 \le E_0, \label{P2-b}\\
& 1 < \tilde b \le B_{\max}, \label{P2-c}\\
& (30d)~{\rm and}~(30e) \nonumber.
\end{align}
\end{subequations}
Problem (P2) remains non-convex due to the non-convex objective and the coupled terms in constraints \eqref{P2-a} and \eqref{P2-b}.

Next, we propose an effective algorithm to solve problem
(P2). To deal with the non-convex constraints in \eqref{P2-a} and \eqref{P2-b}, we first introduce an auxiliary variable $\tilde b' > 0$ and rewrite the constraints using $\tilde b'$. Specifically, (P2) is  reformulated as
\begin{subequations}\label{P3}
\begin{align}
\text{(P3)}:\ \min_{\tilde b,\tilde b',f,\tilde f}\quad 
& \frac{1}{2\lambda}\!\left(\sqrt{1+\frac{4}{2^{\tilde b-1}-1}}-1\right)
 - \frac{1}{\lambda 2^{\tilde b}} \nonumber\\
\text{s.t.}\quad 
& \frac{N_{\rm FLOP}}{\tilde b' b f c}+ \frac{\tilde N_{\rm FLOP}}{\tilde f \tilde c} \le T_0, \label{P3-a}\\
& \eta \frac{N_{\rm FLOP}}{\tilde b' b c}\psi f^2 
+ \tilde\eta \frac{\tilde N_{\rm FLOP}}{\tilde c}\tilde\psi \tilde f^2 \le E_0, \label{P3-b}\\
& \tilde b \le \frac{1}{\tilde b'}, \label{P3-c}\\
& \tilde b'>0, \label{P3-d}\\
& (31c), (30d), {\rm and}~(30e) \nonumber.
\end{align}
\end{subequations}
In problem (P3), constraints \eqref{P3-a} and \eqref{P3-b} are
convex with respect to (w.r.t.) ${\tilde b}^{\prime}$, $f$, and $\tilde f$. However, Problem (P3) is still
non-convex considering the objective function and the non-convex
constraint \eqref{P3-c}.

To address them, we use the successive convex approximation (SCA) technique by approximating the non-convex terms with the first-order Taylor expansion in an iterative manner. First, we focus on the objective function. The second term of the objective function is convex w.r.t. ${\tilde b}$. To tackle it, we approximate its lower bound at a local point ${\tilde b}^{(k)}$. Then, for each iteration $k \ge 1$, the second term of the objective function is lower bounded as 
\begin{align} \label{objective_second_linear}
\frac{1}{\lambda 2^{{\tilde b}}} \ge \frac{1}{\lambda 2^{{\tilde b}^{(k)}}} - \frac{\ln 2}{\lambda 2^{{\tilde b}^{(k)}}} ({\tilde b}-{\tilde b}^{(k)}) \triangleq \underline{\zeta}^{(k)}(\tilde b).
\end{align}
Substituting \eqref{objective_second_linear} into the objective yields the convex upper bound as
\begin{align} \label{objective_upper}
&\frac{1}{2 \lambda}\bigg( \sqrt{1+\frac{4}{2^{\tilde b-1}-1}} -1\bigg)- \frac{1}{\lambda 2^{{\tilde b}}} \nonumber \\
&\le \frac{1}{2 \lambda}\bigg( \sqrt{1+\frac{4}{2^{\tilde b-1}-1}} -1\bigg)- \underline{\zeta}^{(k)}(\tilde b) \triangleq \bar{\zeta}^{(k)}(\tilde b).
\end{align}

Next, we deal with the non-convex constraint \eqref{P3-c}. Also, we approximate \eqref{P3-c} at the local point ${\tilde b}^{\prime (k)}$ based on its first-order Taylor expansion as,  
\begin{align} 
\tilde b - \frac{1}{\tilde b^{\prime}} \le \tilde b - \frac{1}{\tilde b^{\prime^{(k)}}}+\frac{1}{\tilde b^{\prime^{({k})2}}}(\tilde b^{\prime}-\tilde b^{\prime^{(k)}}) \le 0. \label{P2capproximation}
\end{align}

Finally, for the $k$-th iteration, we obtain the approximate convex version of problem (P2) as problem (P4.$k$), via using $\bar{\zeta}^{(k)}(\tilde b)$ in \eqref{objective_upper} as the objective function, and replacing \eqref{P3-c}  with \eqref{P2capproximation}. The convex subproblem at iteration $k$ is given by
\begin{subequations}\label{P4}
	\begin{align}
	\text{(P4.{\it k})}:\mathop {\min }\limits_{\tilde b,{\tilde b}^{\prime},f, \tilde f} & \bar{\zeta}^{(k)}(\tilde b)  \nonumber\\
	\mathrm{s.t.}~&
	\tilde b - \frac{1}{\tilde b^{\prime^{(k)}}}+\frac{1}{\tilde b^{\prime^{({k})2}}}(\tilde b^{\prime}-\tilde b^{\prime^{(k)}}) \le 0 \label{P4-c}\\
	~
	& (32a), (32b), (32d), (31c), (30d), {\rm and}~(30e). \nonumber
	\end{align}
	\end{subequations}

Problem (P4.$k$) can be efficiently solved using standard convex optimization tools (e.g., CVX). By iteratively solving (P4.$k$) and updating $\tilde b^{(k)}$ and $\tilde b'^{(k)}$, the objective value of problem (P2) is monotonically non-increasing. This guarantees the convergence of the proposed SCA-based solution. The iterations terminate when the decrease of the objective value  is below a threshold. Finally, after obtaining the relaxed solution $\tilde b^{\star}$, we recover an achievable bit-width by rounding $\tilde b^{\star}$ to the nearest feasible value in $\mathcal{B}$. The overall algorithm is summarized in Algorithm 1.

\begin{figure*}[h]
    \centering 
	\subfigure[FCDNN-16]{\includegraphics[width=0.3\linewidth]{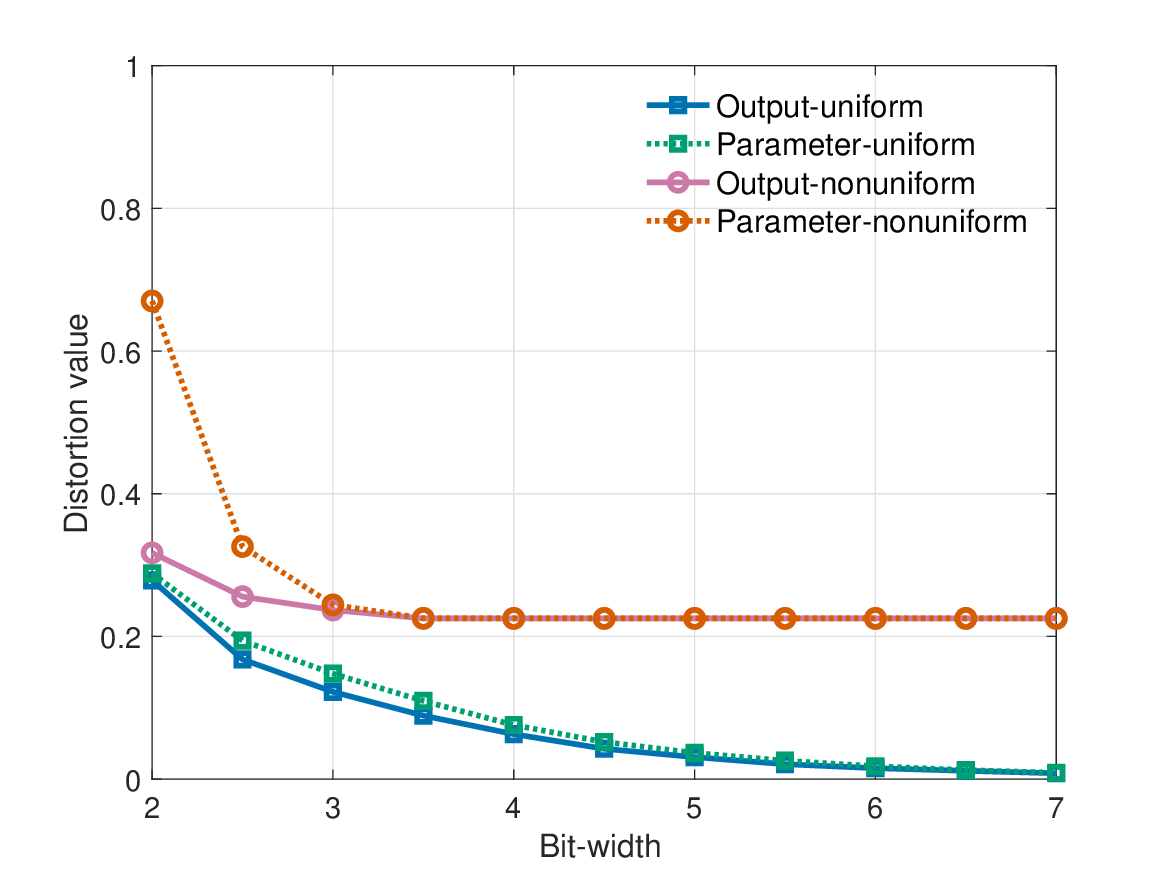}\label{fig:distortion_fcdnn8}}
        \subfigure[BLIP-2]{\includegraphics[width=0.3\linewidth]{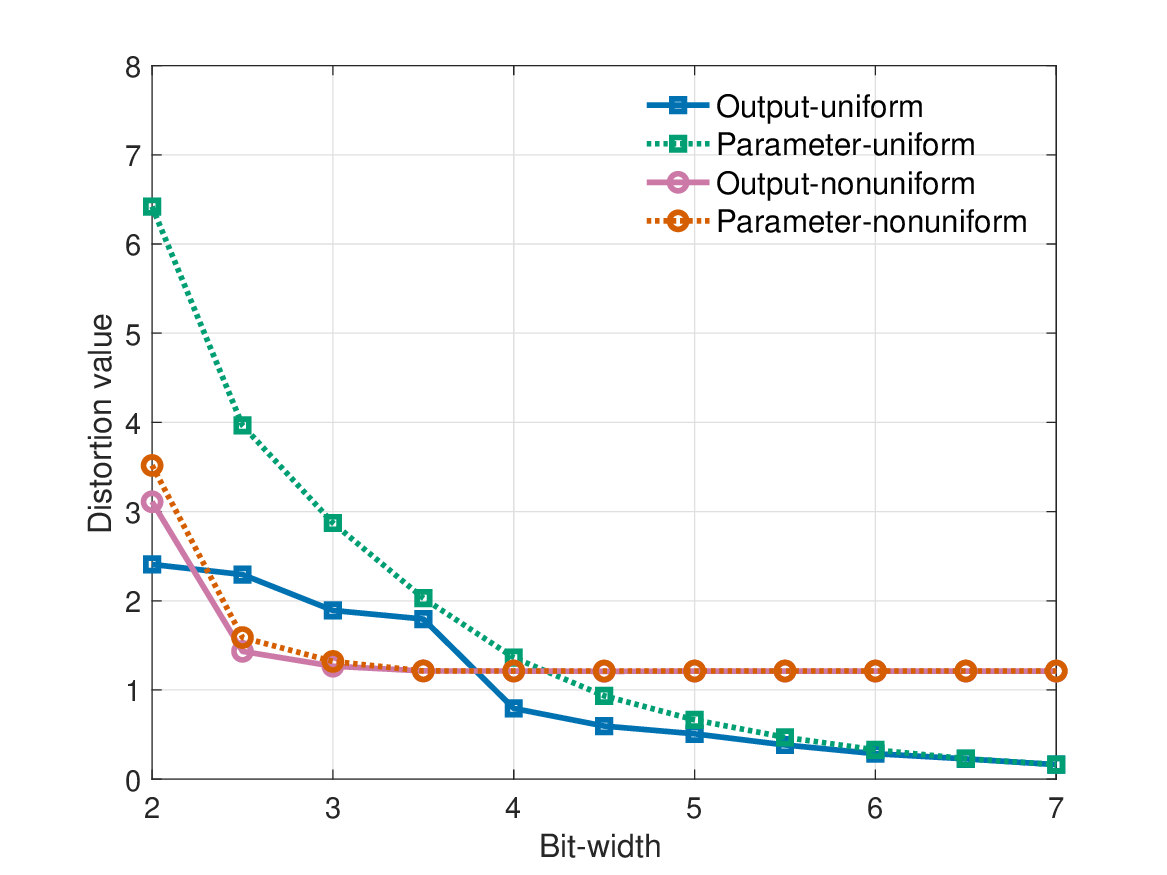}\label{fig:distortion_bert}}
		\subfigure[GIT]{\includegraphics[width=0.3\linewidth]{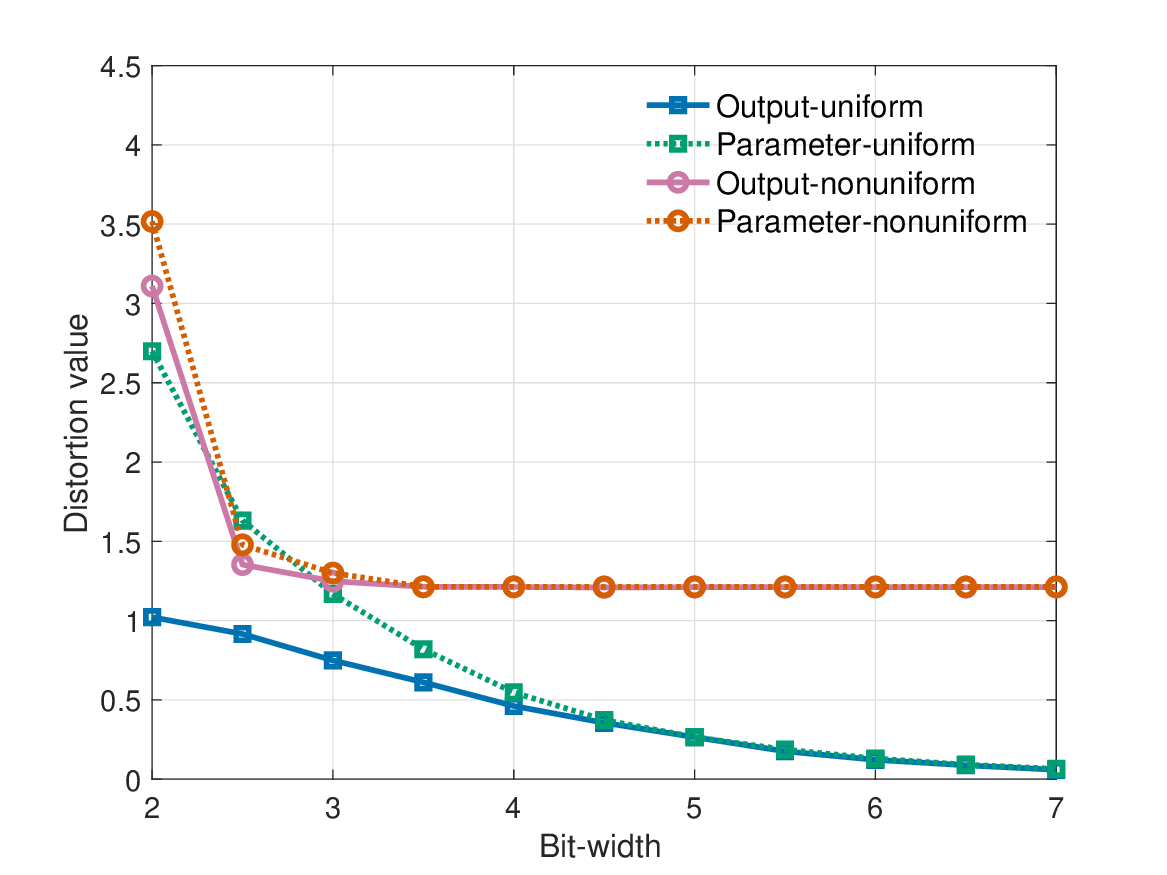}\label{fig:distortion_bart}}
    \caption{Distortions of model outputs and parameters w.r.t. bit-width.}\label{fig:distortion}
\end{figure*}

\section{Evaluation}
We present performance evaluation results to validate the proposed joint quantization and computation design for LAIM co-inference.

\subsection{Verification of Model Output Distortion Approximation}
To validate the model output distortion approximation derived in Section~III, we compare the parameter distortion bounds with the output distortion. Specifically, for each quantization bit-width, we compute the parameter distortion induced by quantization and evaluate the corresponding theoretical distortion upper bound, which is then compared against the actual output distortion. The model-dependent coefficient that relates parameter distortion to output distortion is estimated in a data-driven manner as an empirical upper-bound constant.

We begin by verifying the proposed distortion approximation on an FC DNN with 16 hidden layers, denoted as FCDNN-16. Then  we extend the validation to BLIP-2 \cite{10.5555/3618408.3619222} and GIT \cite{wang2022git}. These models are selected not only because they are widely adopted, but also because their architectures are particularly well suited for co-inference. In the following, we introduce the models in detail.
\begin{itemize}
\item \textbf{FCDNN-16}:  
FCDNN-16 is an autoencoder with ReLU activation functions. The encoder consists of 8 layers with dimensions $[64, 128, 256, 512, 256, 128, 64, 32]$, and the decoder is symmetric. The network is trained using the mean squared error (MSE) loss on MNIST.
\item \textbf{BLIP-2}: We adopt the BLIP-2-2.7b version \cite{10.5555/3618408.3619222} for image understanding. BLIP-2 consists of a frozen CLIP-like vision encoder, a Querying Transformer (Q-Former), and a frozen large language model (LLM) (OPT-2.7B). Such a decoupled modular architecture naturally enables model partitioning across device and edge/cloud servers, making BLIP-2 particularly suitable for co-inference scenarios. BLIP-2-2.7b contains 3.75 billion parameters, and requires 533.66 GFLOPs to generate the first token.
\item \textbf{GIT}: We adopt the Microsoft's GIT-base version \cite{wang2022git} for video understanding. GIT has a two-part architecture with a visual encoder and a text decoder. Similar to BLIP-2, its visual backbone and language generation modules can be cleanly decoupled to allow co-inference. GIT-base contains 176.62 million parameters, and requires 212.27 GFLOPs to generate the first token during inference.
\end{itemize}

To evaluate the robustness of the proposed distortion approximation under different quantization strategies, we consider both uniform and nonuniform quantization schemes. For uniform quantization, model parameters are quantized using a fixed step size, resulting in evenly spaced quantization levels, which is widely adopted due to its simplicity and hardware friendliness \cite{10.5555/3045390.3045690}. For nonuniform quantization, we adopt a widely-used power-of-two logarithmic (PoT-log) quantization scheme \cite{zhou2017incremental}, in which quantization levels are distributed logarithmically.

Fig.~\ref{fig:distortion} shows the measured model output distortion and the theoretical parameter distortion bound under different quantization bit-widths.
It is observed that, across all evaluated models and quantization schemes, the parameter distortion consistently upper bounds the output distortion. 
It is also observed that, as the quantization bit-width increases, the gap between the theoretical distortion bound and the measured distortions gradually narrows. In particular, when the bit-width exceeds a moderate threshold, the theoretical bound becomes notably tight, for example, when  bit-width is larger than 3 for nonuniform, and 4 for uniform quantization.
These results confirm that the proposed model output distortion approximation provides a reliable and computationally tractable characterization of quantization-induced performance distortion, thereby justifying our analysis and its use as a surrogate distortion metric in Sections III and IV.


\begin{figure}[h]
	\centering
	 \epsfxsize=1\linewidth
		\includegraphics[width=7cm]{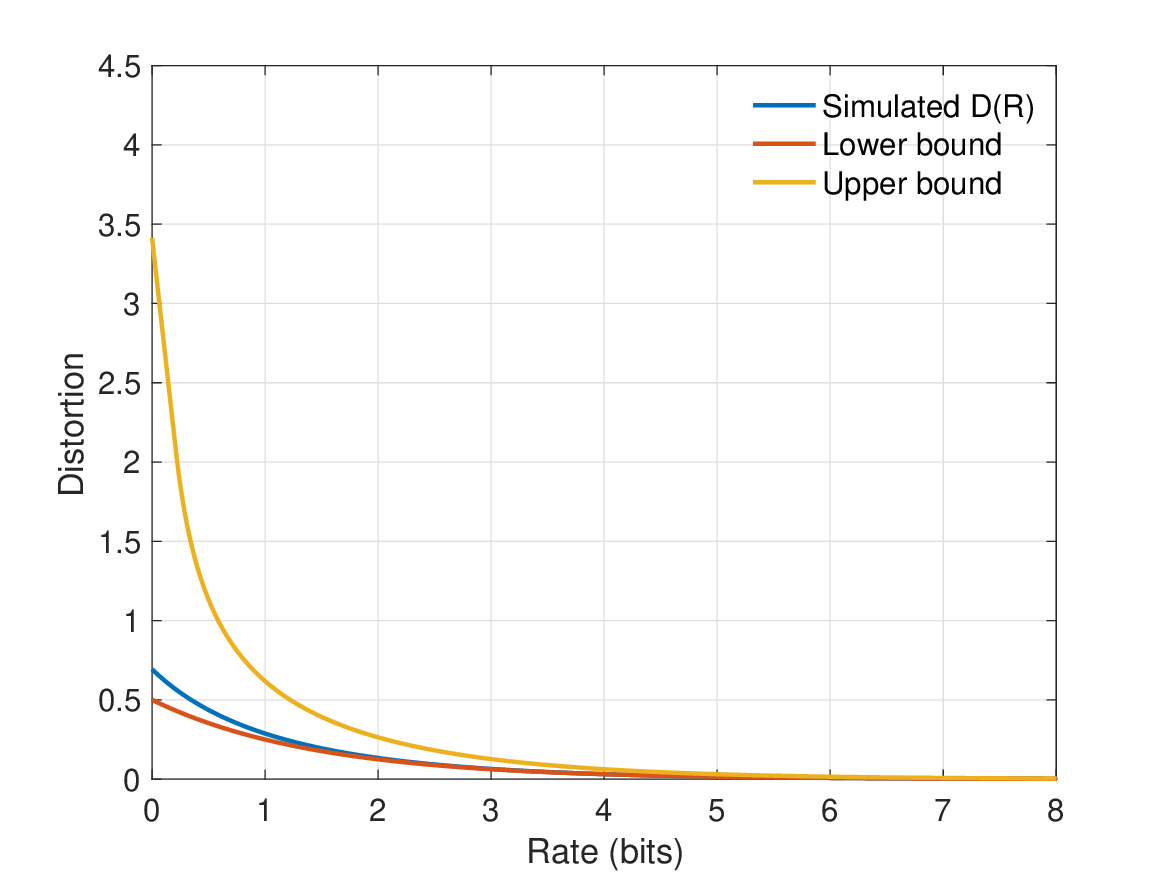}
	\caption{\label{RD}Illustration of the upper and lower  bounds of the distortion-rate function.}
\end{figure}

\begin{figure*}[h]
	\centering
	 \epsfxsize=1\linewidth
		\includegraphics[width=14cm]{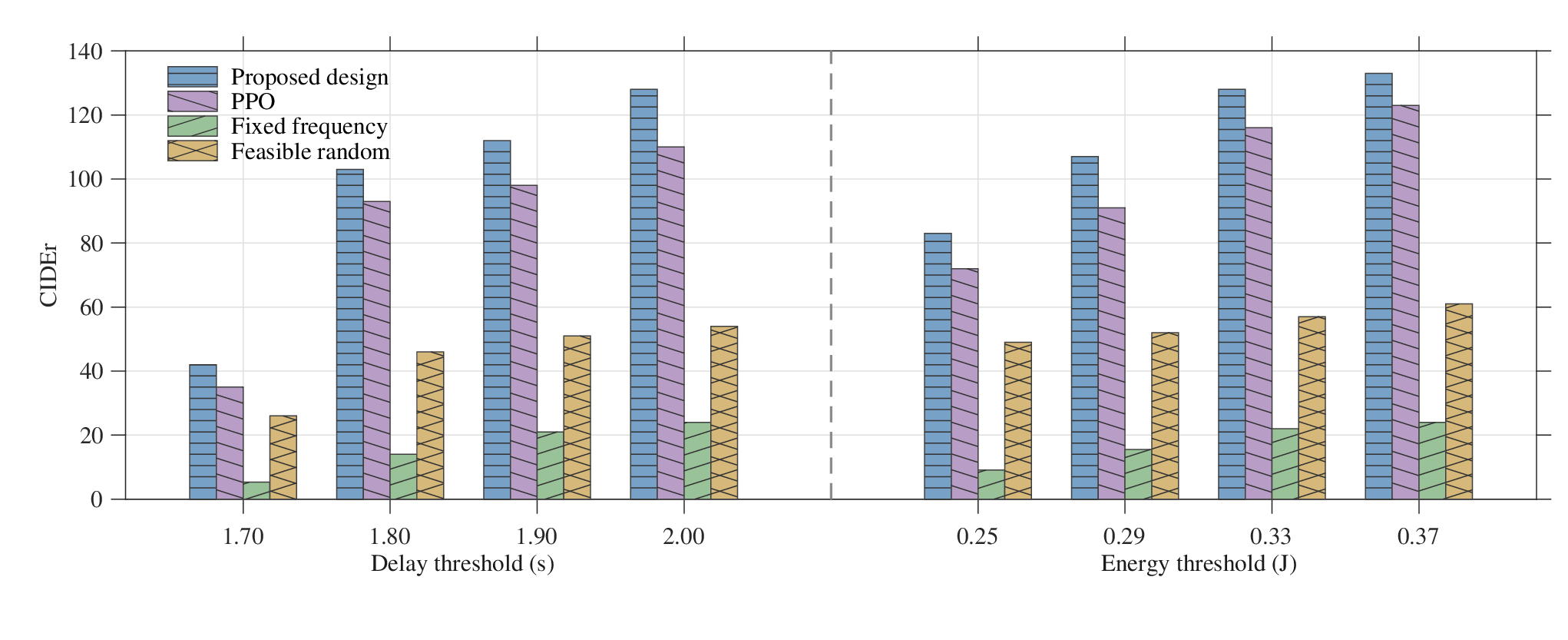}
	\caption{Performance of BLIP-2 on MS-COCO w.r.t. different delay and energy consumption thresholds under uniform quantization with $E_0=2.00~ {\rm J}$ (left) and $T_0= 3.50 ~ {\rm s}$ (right).}\label{BLIP2-uniform}
\end{figure*}

\begin{figure*}[h]
	\centering
	 \epsfxsize=1\linewidth
		\includegraphics[width=14cm]{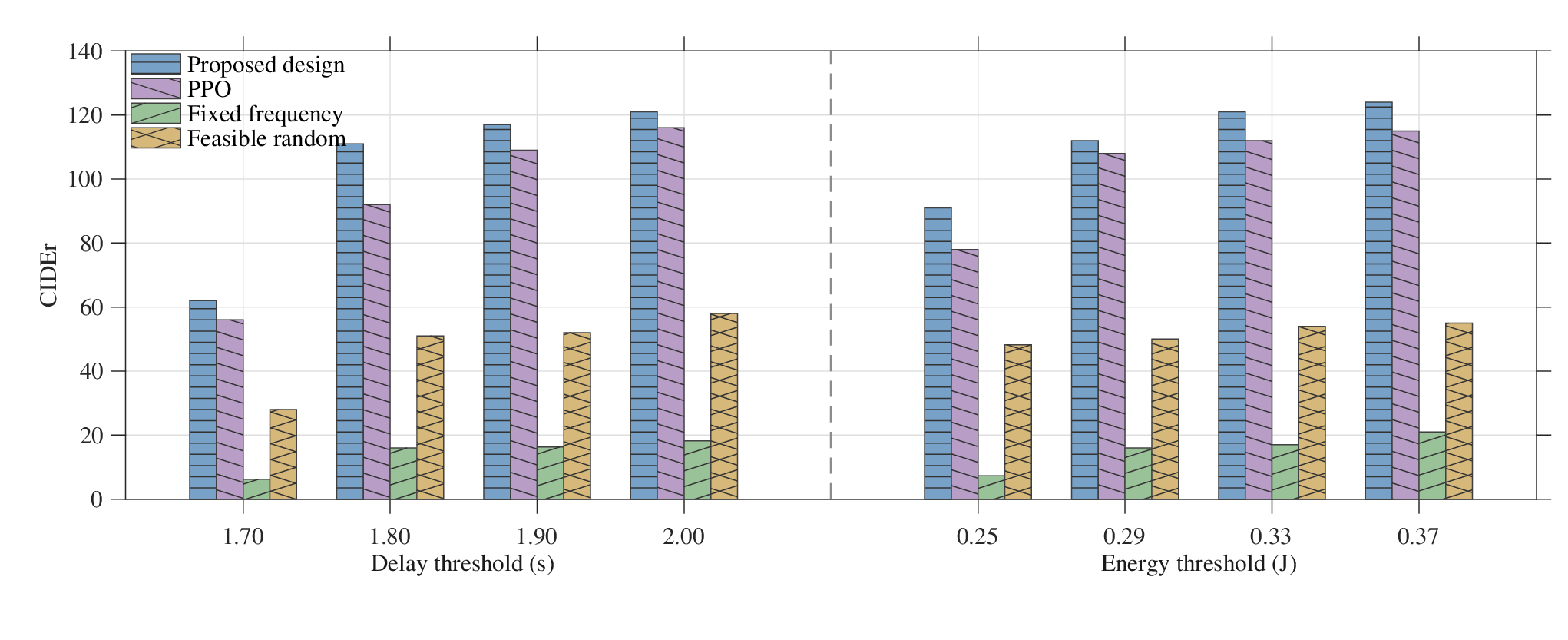}
	\caption{Performance of BLIP-2 on MS-COCO w.r.t. different delay and energy consumption thresholds under nonuniform quantization with $E_0=2.00~ {\rm J}$ (left) and $T_0= 3.50 ~ {\rm s}$ (right).}\label{BLIP2-nonuniform}
\end{figure*}

\subsection{Verification of the Distortion-rate Bounds}

Next, we evaluate the derived distortion-rate bounds by comparing them with a numerically estimated distortion-rate function. Specifically, we plot the analytical lower bound $D^{\rm L}(R)$ in \eqref{eq:SLB-exp-DR} and the upper bound $D^{\rm U}(R)$ in \eqref{eq:UB-exp-DR}, and compare them with a numerically simulated $D(R)$. Since obtaining the exact  $D(R)$ in closed forms is generally difficult for our considered setup, we estimate it numerically using a classical Blahut-Arimoto type procedure \cite{cover1999elements}. Particularly, the continuous source is first approximated by a sufficiently fine discrete alphabet, and the resulting discrete rate-distortion problem is solved via the iterative optimization of the test channel under an average distortion constraint. By sweeping the associated Lagrange multiplier, we obtain a numerical estimate of $D(R)$.

Fig.~\ref{RD} compares the numerically simulated $D(R)$ with the analytical bounds $D^{\rm L}(R)$ and $D^{\rm U}(R)$. It is observed that the simulated $D(R)$ decreases rapidly with the rate, exhibiting an approximately exponential decay, which is consistent with the functional trend predicted by the derived bounds. Next, $D^{\rm U}(R)$ is relatively loose in the very low-rate regime, which is expected since it is constructed by a specific test channel and thus may be conservative when the quantization budget is extremely limited. Nevertheless, as the rate increases to a moderate range (e.g., larger than 2 bits), the gap between $D^{\rm U}(R)$ and $D(R)$ becomes increasingly tight. Meanwhile, $D^{\rm L}(R)$ captures the correct scaling law of distortion reduction w.r.t. the rate (bit-width). Finally, as the rate further increases, both bounds converge toward a small distortion floor, showing that the interval $[D^{\rm L}(R), D^{\rm U}(R)]$ provides an efficient characterization of the distortion-rate behavior for LAIM quantization, which is well suited for guiding subsequent quantization-aware system design.

\begin{figure*}[h]
	\centering
	 \epsfxsize=1\linewidth
		\includegraphics[width=14cm]{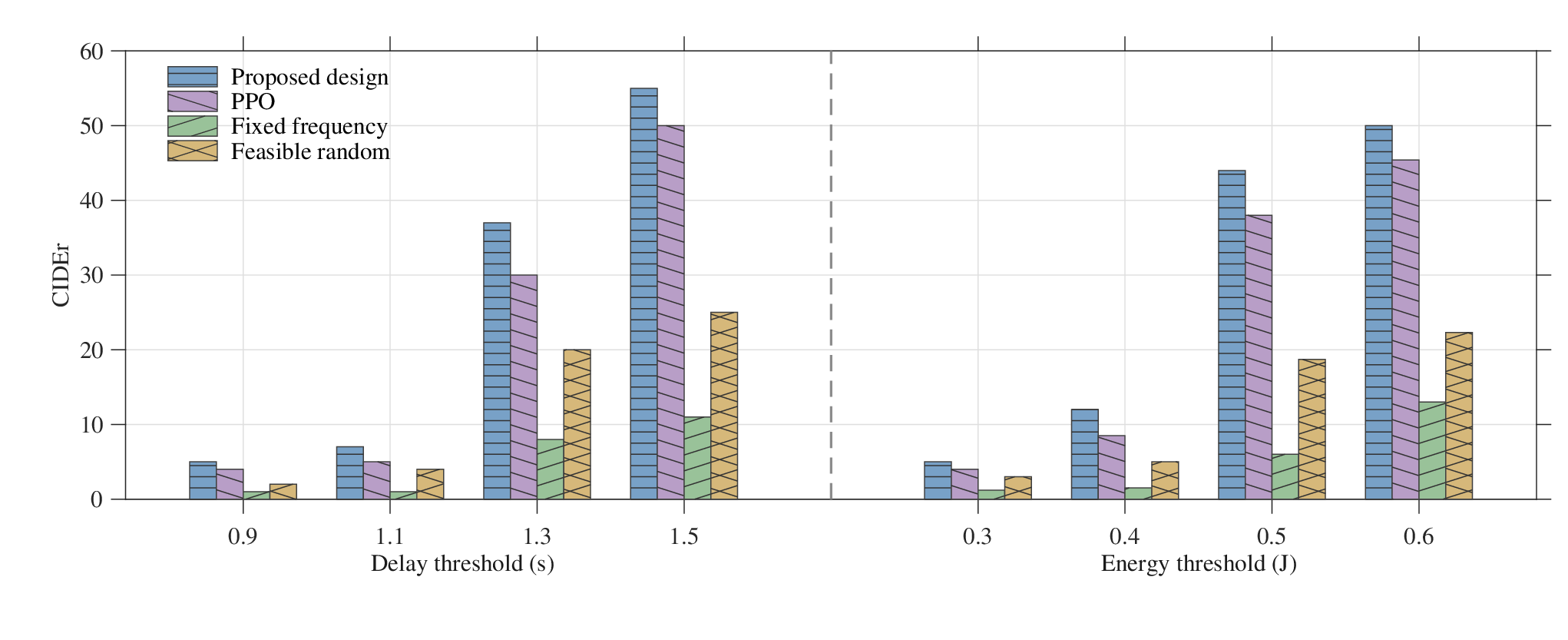}
	\caption{Performance of GIT on VaTeX w.r.t. different delay and energy consumption thresholds under uniform quantization with $E_0=2.00~ {\rm J}$ (left) and $T_0= 2.00 ~ {\rm s}$ (right).}\label{GIT-uniform}
\end{figure*}

\begin{figure*}[h]
	\centering
	 \epsfxsize=1\linewidth
		\includegraphics[width=14cm]{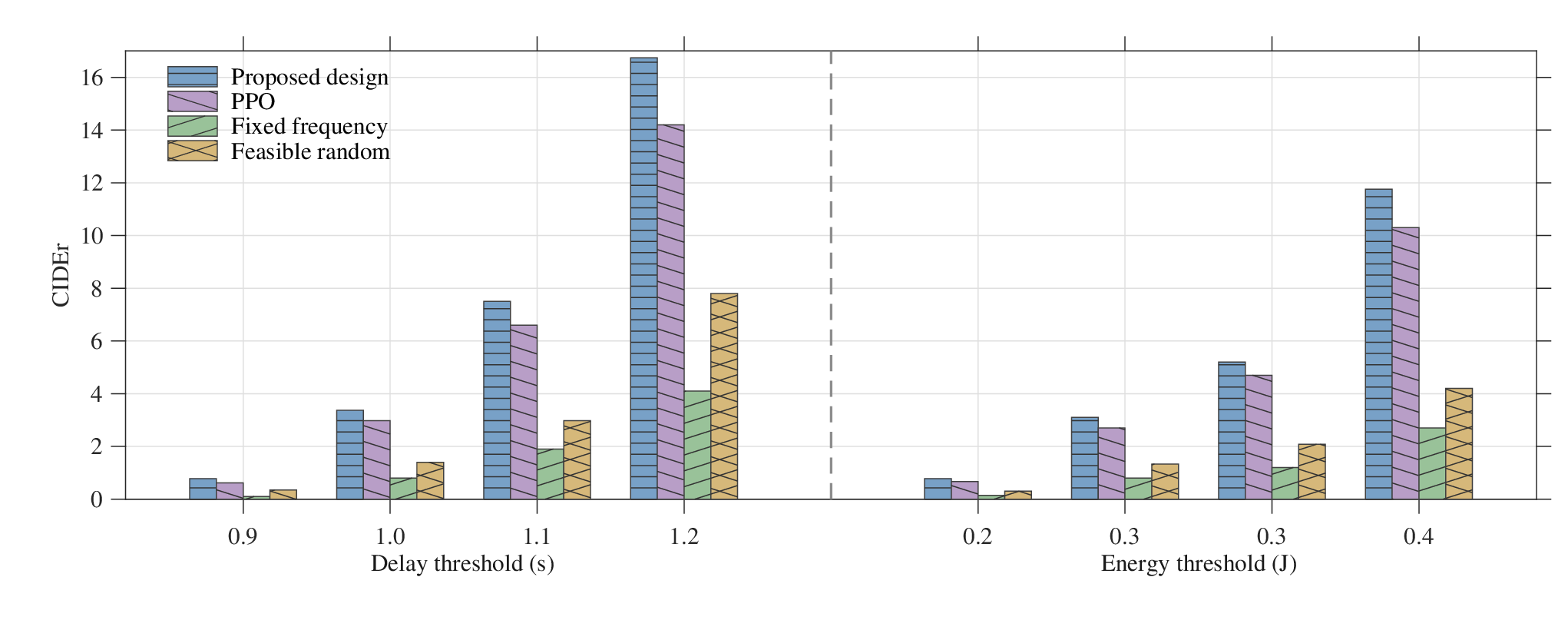}
	\caption{Performance of GIT on VaTeX w.r.t. different delay and energy consumption threshold under nonuniform quantization with $E_0=2.00~ {\rm J}$ (left) and $T_0= 2.00 ~ {\rm s}$ (right).}\label{GIT-nonuniform}
\end{figure*}

\subsection{Performance of the Proposed Joint Quantization and Computation Design}
We evaluate the proposed joint quantization and computation design using two widely adopted vision datasets, i.e., MS-COCO \cite{Lin2014COCO} for image and VaTeX \cite{Wang2019VaTeX} for video understanding. Specifically,
\begin{itemize}
	\item MS-COCO is a large-scale image dataset of complex everyday scenes with diverse resolutions and rich object contexts. For image captioning, it provides five independent human-written reference captions for each image. Following the widely used Karpathy split for evaluation, the dataset is partitioned into $113,287$ images for training, and $5,000$ images each for validation and testing.
	\item VaTeX is a large-scale video-language dataset consisting of around 26k, 3k, and 6k video clips for training, validation, and testing, with each clip lasting around 10 seconds. We uniformly sample four frames from each video clip along the temporal axis as the model input.
\end{itemize}
For each dataset, the inference quality is measured by the CIDEr score, a consensus-based metric that computes the similarity between a generated sentence and a set of ground-truth references using weighted $n$-gram matching. Specifically, given a candidate sentence $p_i$ and a reference set $\hat{P}_i=\{\hat{p}_{ij}\}_{j=1}^{m}$ for sample $i$, we have
\begin{equation}
\mathrm{CIDEr}_n(p_i,\hat{P}_i) = \frac{1}{m}\sum_{j=1}^{m} 
\frac{\mathbf{g}_n(p_i)^{\mathsf{T}}\mathbf{g}_n(\hat{p}_{ij})}{\|\mathbf{g}_n(p_i)\|\,\|\mathbf{g}_n(\hat{p}_{ij})\|},
\end{equation}
where  $\mathbf{g}_n(\cdot)$ is the weighted $n$-gram vectors. We obtain the overall CIDEr score by averaging over different $n$-gram orders.

We consider a similar simulation setup as in \cite{11301737} on 2 Nvidia RTX 3090 GPUs. Specifically, the maximum clock frequencies are set to $f^{\rm max} = 2~\mathrm{GHz}$ and $\tilde{f}^{\rm max} = 10~\mathrm{GHz}$, and the number of FLOPs per cycle are set to $c = 32$ and $\tilde{c} = 128$. The PUEs are set to $\eta = 1$ and $\tilde{\eta} = 2$, and the power coefficients are set to $\psi = 2 \times 10^{-29}~\mathrm{W/(cycle/s)^3}$ and $\tilde{\psi} = 1 \times 10^{-28}~\mathrm{W/(cycle/s)^3}$.
We compare the proposed design with three benchmark schemes: 1) PPO-based design \cite{chen2024adaptive}: This is the state-of-the-art method, where the system is modeled as a Markov decision process. Then the quantization and computation frequencies design problem is solved via DRL. 2) Fixed-frequency design: The device and server frequencies are fixed to predetermined maximum values and only the bit-width is optimized to satisfy the QoS constraints. 3) Feasible random design: We randomly sample bit-widths for 400 trials, and for each sampled bit-width, we check feasibility by optimizing the remaining computation frequency variables. Then, only feasible trials are evaluated and reported.

\begin{table*}[h]
\centering
\caption{Co-inference performance (CIDEr score) evaluated on the developed testbed.}
\label{tab:delay_energy}
\resizebox{0.9\linewidth}{!}{
\begin{tabular}{ccccccccccccc}
\toprule
\multirow{2}{*}{} 
& \multicolumn{6}{c}{BLIP-2} 
& \multicolumn{6}{c}{GIT} \\
\cmidrule(lr){2-7} \cmidrule(lr){8-13}
& \multicolumn{3}{c}{Delay (s)} 
& \multicolumn{3}{c}{Energy (J)} 
& \multicolumn{3}{c}{Delay (s)} 
& \multicolumn{3}{c}{Energy (J)} \\
\cmidrule(lr){2-4} \cmidrule(lr){5-7}
\cmidrule(lr){8-10} \cmidrule(lr){11-13}
& 2.50 & 2.55 & 2.60 
& 164.8 & 165.0 & 165.2 
& 1.38 & 1.40 & 1.42 
& 19.8 & 20.0 & 20.2 \\
\midrule
Low    
& 8.2  & 61.4 & 109.2 
& \textbf{133.4} & \textbf{134.6} & \textbf{138.8}
& 5.4  & 7.5  & 18.2
& \textbf{46.2}  & \textbf{50.4}  & \textbf{53.7} \\
Medium 
& 43.6 & 102.7 & 127.5
& 115.3 & 128.1 & 131.7
& 38.5 & 43.7 & 48.6
& 26.7 & 38.9 & 48.8 \\
High   
& \textbf{100.8} & \textbf{118.9} & \textbf{132.4}
& 8.2 & 43.9 & 101.9
& \textbf{54.3} & \textbf{55.9} & \textbf{57.1}
& 5.8 & 21.3 & 38.6 \\
\bottomrule
\end{tabular}}
\end{table*}

Figs.~\ref{BLIP2-uniform}, \ref{BLIP2-nonuniform}, \ref{GIT-uniform}, and \ref{GIT-nonuniform} illustrate the CIDEr performance of BLIP2 and GIT under different delay and energy constraints, for uniform and nonuniform quantization schemes, respectively. It is observed that the proposed design consistently achieves the highest CIDEr score, demonstrating the effectiveness of the proposed joint quantization and computation design in exploiting the inference quality, delay, and energy consumption trade-off.
The performance advantage over the PPO-based baseline mainly stems from the fact that DRL relies on proper initialization, sufficient exploration, and penalty-driven constraint handling, which may result in suboptimal solutions. 
Compared with the fixed-frequency and feasible random baselines, the performance gain of the proposed design is even more pronounced, as it explicitly exploits a higher design DoF for co-inference through joint optimization of quantization and computation resources.
Moreover, as the delay or energy constraints become more relaxed, the CIDEr score generally increases. This is because a larger delay or energy budget provides more flexibility in system design, allowing higher quantization precision and/or more favorable computation frequencies.
Finally, the proposed design exhibits robust performance across different quantization schemes, indicating that it generalizes well to both uniform and nonuniform quantization.

Finally, we implement real-world experiments using an NVIDIA Jetson AGX Orin 64GB as the edge device and a Dell PowerEdge R740 server equipped with dual Intel Xeon Gold 6246R CPUs and two NVIDIA RTX 3090 GPUs as the edge server. The device and server are connected through a commercial Wi-Fi router operating under stable 5 GHz WLAN networks. The complete testbed setup is illustrated in Fig.~\ref{testbed}.
In the current testbed, it is challenging to precisely control the computation frequency at the device side. To enable effective  evaluation, we adopt three practically accessible operating profiles, namely low-frequency, medium-frequency, and high-frequency configurations, each corresponding to a feasible frequency setting supported by the Jetson AGX Orin. Under each configuration, we evaluate the co-inference performance under scenarios with only the delay or the energy consumption constraint.

Table~\ref{tab:delay_energy} reports the co-inference performance under different delay  and energy consumption constraints. It is observed that, in the delay-limited but energy-sufficient regime, adopting a higher-frequency profile yields better task performance. This is because increasing the computation frequency directly reduces processing latency, thereby allowing the system to employ a larger quantization bit-width without violating the delay constraint. As a result, quantization-induced distortion is reduced, leading to improved inference quality.
In contrast, in the energy-limited but delay-sufficient regime, a lower-frequency profile achieves better task performance. Under a tight energy budget, operating at a higher frequency may increase energy consumption, which forces the system to compensate by adopting a more aggressive quantization scheme to remain feasible, thereby degrading inference quality. 
The results show that, even when fine-grained frequency control is unavailable, a coarse-grained selection among a small set of feasible frequency profiles remains effective. This further demonstrates the necessity and practicality of the proposed joint quantization and computation design for improving LAIM co-inference performance in real edge environments.

\begin{figure}[h]
\centering
\includegraphics[width=0.8\linewidth]{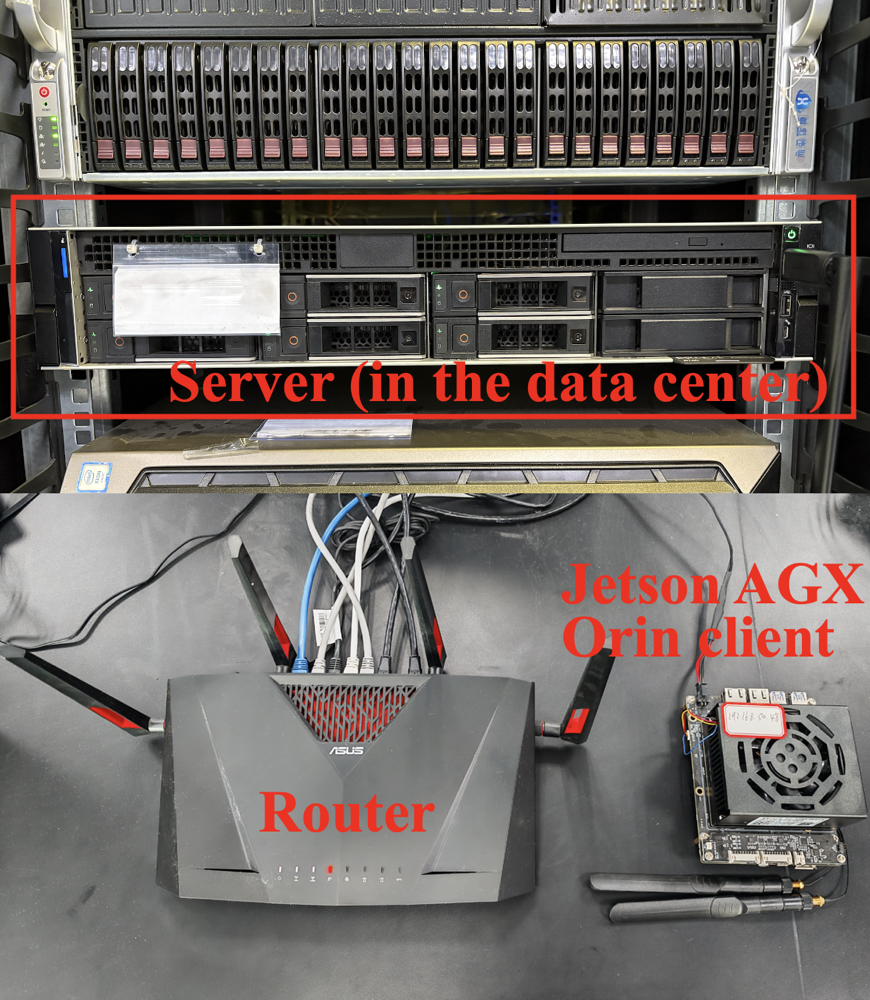}
\caption{\label{testbed} Real-world testbed for quantization-aware LAIM co-inference.}\label{testbed}
\end{figure}

\section{Conclusions}
This paper has investigated quantization-aware LAIM co-inference for embodied AI systems. First, we have established a tractable approximation for quantization-induced inference distortion via parameter-level perturbations, and derived lower and upper bounds on the rate-distortion function to characterize its dependence on LAIM statistics, such as the quantization bit-width. Then, we have formulated a joint quantization and computation design problem under QoS constraints, and developed an efficient solution to minimize the distortion upper bound while maintaining a tight approximation gap via maximizing the lower bound. Extensive simulations and real-world testbed experiments have demonstrated that the proposed design consistently outperforms benchmark schemes in  improving LAIM co-inference performance in embodied AI systems.
\begin{appendix}

\subsection{Proof of Proposition~\ref{proposition1}}

We upper bound the inference output distortion of the FC DNN under model quantization. Specifically, we have
\begin{align}
&\|f(\mv{x},\mv{W}) - f(\mv{x},\hat{\mv{W}})\|_1 \nonumber \\
&= \big\|\mv{W}^{(L)} \sigma\!\big(f(\mv{x},\mv{W}^{(1:L-1)})\big)
      - \hat{\mv{W}}^{(L)} \sigma\!\big(f(\mv{x},\hat{\mv{W}}^{(1:L-1)})\big)\big\|_1 \nonumber \\
&\le \|\mv{W}^{(L)} - \hat{\mv{W}}^{(L)}\|_1
     \big\|\sigma\!\big(f(\mv{x},\mv{W}^{(1:L-1)})\big)\big\|_1 \nonumber \\
&\quad + \|\hat{\mv{W}}^{(L)}\|_1
     \big\|\sigma\!\big(f(\mv{x},\mv{W}^{(1:L-1)})\big)
     - \sigma\!\big(f(\mv{x},\hat{\mv{W}}^{(1:L-1)})\big)\big\|_1, \label{eq:app1}
\end{align}
where the inequality follows from the triangle inequality. By Assumptions 2 and 3, we further obtain
\begin{align}
&\|f(\mv{x},\mv{W}) - f(\mv{x},\hat{\mv{W}})\|_1 \nonumber \\
&\le \|\mv{W}^{(L)} - \hat{\mv{W}}^{(L)}\|_1
     \|f(\mv{x},\mv{W}^{(1:L-1)})\|_1 \nonumber \\
& + (\|\mv{W}^{(L)}\|_1+ \tau^{(L)})
     \|f(\mv{x},\mv{W}^{(1:L-1)})
     - f(\mv{x},\hat{\mv{W}}^{(1:L-1)})\|_1. \label{eq:app2}
\end{align}

Next, we upper bound $\|f(\mv{x},\mv{W}^{(1:L-1)})\|_1$. By recursively applying Assumption 2 and the Cauchy-Schwarz inequality, we have
\begin{align}
\|f(\mv{x},\mv{W}^{(1:L-1)})\|_1
&\le \|\mv{W}^{(L-1)}\|_1
      \big\|\sigma\!\big(f(\mv{x},\mv{W}^{(1:L-2)})\big)\big\|_1 \nonumber \\
&\le \|\mv{W}^{(L-1)}\|_1
      \|f(\mv{x},\mv{W}^{(1:L-2)})\|_1 \nonumber \\
&\le \prod_{j=1}^{L-1} \|\mv{W}^{(j)}\|_1. \label{eq:app3}
\end{align}
Substituting \eqref{eq:app3} into \eqref{eq:app2} yields
\begin{align}
&\|f(\mv{x},\mv{W}) - f(\mv{x},\hat{\mv{W}})\|_1 \nonumber \\
&\le \|\mv{W}^{(L)} - \hat{\mv{W}}^{(L)}\|_1
      \prod_{j=1}^{L-1} \|\mv{W}^{(j)}\|_1 \nonumber \\
& + (\|\mv{W}^{(L)}\|_1 + \tau^{(L)})
      \|f(\mv{x},\mv{W}^{(1:L-1)})
      - f(\mv{x},\hat{\mv{W}}^{(1:L-1)})\|_1. \label{eq:app4}
\end{align}
Dividing both sides of \eqref{eq:app4} by $\prod_{j=1}^{L} \|\mv{W}^{(j)}\|_1$, we obtain
\begin{align}
&\frac{\|f(\mv{x},\mv{W}) - f(\mv{x},\hat{\mv{W}})\|_1}
      {\prod_{j=1}^{L} \|\mv{W}^{(j)}\|_1} \nonumber \\
&\le \frac{\|\mv{W}^{(L)} - \hat{\mv{W}}^{(L)}\|_1}
           {\|\mv{W}^{(L)}\|_1} \nonumber \\
&+ \Big(1 + \frac{\tau^{(L)}}{\|\mv{W}^{(L)}\|_1}\Big)
      \frac{\|f(\mv{x},\mv{W}^{(1:L-1)})
      - f(\mv{x},\hat{\mv{W}}^{(1:L-1)})\|_1}
      {\prod_{j=1}^{L-1} \|\mv{W}^{(j)}\|_1}. \label{eq:app5}
\end{align}
Then, we apply the same procedure on the second term in \eqref{eq:app5} and obtain
\begin{align}
&\frac{\|f(\mv{x},\mv{W}) - f(\mv{x},\hat{\mv{W}})\|_1}
      {\prod_{j=1}^{L} \|\mv{W}^{(j)}\|_1} \nonumber \\
&\le \sum_{i=1}^{L}
     \prod_{j=i+1}^{L}
     \Big(1 + \frac{\tau^{(j)}}{\|\mv{W}^{(j)}\|_1}\Big)
     \frac{\|\mv{W}^{(i)} - \hat{\mv{W}}^{(i)}\|_1}
          {\|\mv{W}^{(i)}\|_1}. \label{eq:app6}
\end{align}

Finally, by multiplying both sides of \eqref{eq:app6} by $\prod_{j=1}^{L} \|\mv{W}^{(j)}\|_1$, we have
\begin{align}
&\|f(\mv{x},\mv{W}) - f(\mv{x},\hat{\mv{W}})\|_1 \le \nonumber \\
&\sum_{i=1}^{L}\!\!
     \Bigg(
     \!\prod_{j=1}^{i-1} \|\mv{W}^{(j)}\|_1
     \!\!\!\!\prod_{k=i+1}^{L} (\|\mv{W}^{(k)}\|_1 + \tau^{(k)})
     \Bigg)
     \|\mv{W}^{(i)} \!-\!\hat{\mv{W}}^{(i)}\|_1,
\end{align}
which completes the proof.

\subsection{Proof of Lemma~\ref{lemma2}}

We consider the following entropy maximization problem over all PDFs $P_Z(z)$ supported on $\mathbb{R}$,
\begin{align}
\max_{P_Z(z)} \quad 
& h(P_Z(z)) \triangleq -\int_{\mathbb{R}} P_Z(z)\log_2 P_Z(z)\,dz \nonumber\\
\text{s.t.}\quad
& \int_{\mathbb{R}} P_Z(z)\,dz = 1, \label{eq:maxent_norm}\\
& \int_{\mathbb{R}} |z| P_Z(z)\,dz \le D, \label{eq:maxent_moment}\\
& P_Z(z)\ge 0,\ \forall z\in\mathbb{R}. \nonumber
\end{align}

First the constraint \eqref{eq:maxent_moment} must be active at the optimum. Indeed, if a feasible PDF $P_Z(z)$ satisfies $\int |z|P_Z(z) dz < D$, then scaling the random variable as $z' = a z$ with some $a>1$ strictly increases the differential entropy while still satisfying the constraint, which contradicts optimality. Hence, the optimal solution satisfies $\int |z| P_Z(z) dz = D$.

Since the problem is convex and Slater's condition holds, we invoke the Karush-Kuhn-Tucker (KKT) conditions to solve it. We introduce Lagrange multipliers $\lambda_0 \in \mathbb{R}$ and $\lambda_1 \ge 0$ corresponding to constraints \eqref{eq:maxent_norm} and \eqref{eq:maxent_moment}, respectively. The Lagrangian is given by
\begin{align}
\mathcal{L}(P;\lambda_0,\lambda_1)
&\!=\! \!-\!\int_{\mathbb{R}}\!\! P_Z(z)\log_2 P_Z(z)\,dz
\!+\!\lambda_0\!\!\left(\int_{\mathbb{R}} P_Z(z)\,dz - 1\right) \nonumber\\
&\quad +\lambda_1\!\left(D - \int_{\mathbb{R}} |z| P_Z(z)\,dz\right).
\label{eq:Lagrangian}
\end{align}
Let $P_Z^\star(z)$ denote the optimal solution. Consider any admissible perturbation $\eta(z)$ satisfying $\int_{\mathbb{R}} \eta(z)\,dz = 0$, and define $P_\epsilon(z)=P_Z^\star(z)+\epsilon\,\eta(z)$ for sufficiently small $\epsilon$. The stationarity condition requires that the derivative of $\mathcal{L}$ at $\epsilon=0$ equals to 0, i.e., 
\begin{align}
0
&=\frac{d}{d\epsilon}\mathcal{L}(P_\epsilon;\lambda_0,\lambda_1)\Big|_{\epsilon=0} \nonumber\\
&=\int_{\mathbb{R}} \eta(z)\Big(-\big(1+\log_2 P_Z^\star(z)\big)
+\lambda_0-\lambda_1|z|\Big) dz.
\label{eq:variation0}
\end{align}
Solving  \eqref{eq:variation0} yields the exponential-form solution
\begin{align}
P_Z^\star(z)=C e^{-\lambda_1|z|},
\label{eq:exp_form}
\end{align}
where $C = 2^{\lambda_0-1}>0$ is a normalization constant.
Finally, enforcing constraints \eqref{eq:maxent_norm}  and \eqref{eq:maxent_moment} yields the optimal PDF as 
\begin{align}
P^\star(z)=\frac{1}{2D}e^{-|z|/D},
\end{align}
which completes the proof.

\end{appendix}

\bibliographystyle{IEEEtran}
\bibliography{reference}

\begin{thebibliography}{10}
\providecommand{\url}[1]{#1}
\csname url@samestyle\endcsname
\providecommand{\newblock}{\relax}
\providecommand{\bibinfo}[2]{#2}
\providecommand{\BIBentrySTDinterwordspacing}{\spaceskip=0pt\relax}
\providecommand{\BIBentryALTinterwordstretchfactor}{4}
\providecommand{\BIBentryALTinterwordspacing}{\spaceskip=\fontdimen2\font plus
\BIBentryALTinterwordstretchfactor\fontdimen3\font minus
  \fontdimen4\font\relax}
\providecommand{\BIBforeignlanguage}[2]{{%
\expandafter\ifx\csname l@#1\endcsname\relax
\typeout{** WARNING: IEEEtran.bst: No hyphenation pattern has been}%
\typeout{** loaded for the language `#1'. Using the pattern for}%
\typeout{** the default language instead.}%
\else
\language=\csname l@#1\endcsname
\fi
#2}}
\providecommand{\BIBdecl}{\relax}
\BIBdecl

\bibitem{kundu2025airantransformingranaidriven}
\BIBentryALTinterwordspacing
L.~Kundu, X.~Lin, R.~Gadiyar, J.-F. Lacasse, and S.~Chowdhury, ``{AI-RAN:
  Transforming RAN with AI-driven computing infrastructure},'' 2025. [Online].
  Available: \url{https://arxiv.org/abs/2501.09007}
\BIBentrySTDinterwordspacing

\bibitem{10791415}
Z.~Lyu, Y.~Li, G.~Zhu, J.~Xu, H.~Vincent~Poor, and S.~Cui, ``Rethinking
  resource management in edge learning: A joint pre-training and fine-tuning
  design paradigm,'' \emph{IEEE Trans. Wireless Commun.}, vol.~24, no.~2, pp.
  1584--1601, Feb. 2025.

\bibitem{gupta2021embodied}
A.~Gupta, S.~Savarese, S.~Ganguli, and L.~Fei-Fei, ``Embodied intelligence via
  learning and evolution,'' \emph{Nat. Commun.}, vol.~12, no.~1, p. 5721, Oct.
  2021.

\bibitem{10648594}
M.~Xu, D.~Niyato, J.~Kang, Z.~Xiong, S.~Mao, Z.~Han, D.~I. Kim, and K.~B.
  Letaief, ``When large language model agents meet {6G} networks: Perception,
  grounding, and alignment,'' \emph{IEEE Wireless Commun.}, vol.~31, no.~6, pp.
  63--71, Dec. 2024.

\bibitem{10906629}
R.~Yi, L.~Guo, S.~Wei, A.~Zhou, S.~Wang, and M.~Xu, ``{EdgeMoE}: Empowering
  sparse large language models on mobile devices,'' \emph{IEEE Trans. Mob.
  Comput.}, vol.~24, no.~8, pp. 7059--7073, Aug. 2025.

\bibitem{10.1145/3666025.3699355}
Y.~Zhuang, Z.~Zheng, F.~Wu, and G.~Chen, ``{LiteMoE}: Customizing on-device
  {LLM} serving via proxy submodel tuning,'' in \emph{Proc. ACM Int. Conf.
  Embedded Artificial Intelligence and Sensing Systems}, 2024, pp. 521--534.

\bibitem{griggs2024melangecostefficientlarge}
\BIBentryALTinterwordspacing
T.~Griggs, X.~Liu, J.~Yu, D.~Kim, W.-L. Chiang, A.~Cheung, and I.~Stoica,
  ``M\'elange: Cost efficient large language model serving by exploiting {GPU}
  heterogeneity,'' 2024. [Online]. Available:
  \url{https://arxiv.org/abs/2404.14527}
\BIBentrySTDinterwordspacing

\bibitem{jiang2025thunderservehighperformancecostefficientllm}
\BIBentryALTinterwordspacing
Y.~Jiang, F.~Fu, X.~Yao, T.~Wang, B.~Cui, A.~Klimovic, and E.~Yoneki,
  ``Thunderserve: High-performance and cost-efficient {LLM} serving in cloud
  environments,'' 2025. [Online]. Available:
  \url{https://arxiv.org/abs/2502.09334}
\BIBentrySTDinterwordspacing

\bibitem{10.1145/3676641.3716025}
J.~Stojkovic, C.~Zhang, I.~Goiri, E.~Choukse, H.~Qiu, R.~Fonseca, J.~Torrellas,
  and R.~Bianchini, ``Tapas: Thermal- and power-aware scheduling for {LLM}
  inference in cloud platforms,'' in \emph{Proc. ACM Int. Conf. Architectural
  Support for Programming Languages and Operating Systems}, 2025, pp.
  1266--1281.

\bibitem{10591707}
Y.~He, J.~Fang, F.~R. Yu, and V.~C. Leung, ``Large language models ({LLMs})
  inference offloading and resource allocation in cloud-edge computing: An
  active inference approach,'' \emph{IEEE Trans. Mob. Comput.}, vol.~23,
  no.~12, pp. 11\,253--11\,264, Dec. 2024.

\bibitem{10.1145/3704413.3764429}
L.~Yuan, D.-J. Han, S.~Wang, and C.~Brinton, ``Local-cloud inference offloading
  for {LLMs} in multi-modal, multi-task, multi-dialogue settings,'' in
  \emph{Proc. International Symposium on Theory, Algorithmic Foundations, and
  Protocol Design for Mobile Networks and Mobile Computing}, 2025, pp.
  201--210.

\bibitem{chen2024adaptive}
Y.~Chen, R.~Li, X.~Yu, Z.~Zhao, and H.~Zhang, ``Adaptive layer splitting for
  wireless {LLM} inference in edge computing: A model-based reinforcement
  learning approach,'' \emph{Front. Inform. Technol. Electron. Eng.}, vol.~26,
  pp. 278--292, Mar. 2025.

\bibitem{11140540}
S.~Oh, J.~Kim, J.~Park, S.-W. Ko, T.~Q.~S. Quek, and S.-L. Kim,
  ``Uncertainty-aware hybrid inference with on-device small and remote large
  language models,'' in \emph{Proc. IEEE Int. Conf. Machine Learning in
  Communications and Networking}, 2025, pp. 1--7.

\bibitem{10818760}
M.~Zhang, X.~Shen, J.~Cao, Z.~Cui, and S.~Jiang, ``Edgeshard: Efficient {LLM}
  inference via collaborative edge computing,'' \emph{IEEE Internet Things J.},
  vol.~12, no.~10, pp. 13\,119--13\,131, May 2025.

\bibitem{11301737}
Z.~Lyu, M.~Xiao, J.~Xu, M.~Skoglund, and M.~D. Renzo, ``The larger the merrier?
  efficient large {AI} model inference in wireless edge networks,'' \emph{IEEE
  J. Sel. Areas Commun.}, pp. 1--1, 2025.

\bibitem{10854360}
X.~Peng, Z.~Qin, X.~Tao, J.~Lu, and K.~B. Letaief, ``A robust image semantic
  communication system with multi-scale vision transformer,'' \emph{IEEE J.
  Sel. Areas Commun.}, vol.~43, no.~4, pp. 1278--1291, Apr. 2025.

\bibitem{jiang2025largeaimodelsagentic}
\BIBentryALTinterwordspacing
F.~Jiang, C.~Pan, L.~Dong, K.~Wang, O.~A. Dobre, and M.~Debbah, ``From large
  {AI} models to agentic {AI}: A tutorial on future intelligent
  communications,'' 2025. [Online]. Available:
  \url{https://arxiv.org/abs/2505.22311}
\BIBentrySTDinterwordspacing

\bibitem{11038757}
J.~Park, Y.~Oh, Y.~Kim, and Y.-S. Jeon, ``Vision transformer-based semantic
  communications with importance-aware quantization,'' \emph{IEEE Internet
  Things J.}, vol.~12, no.~17, pp. 35\,662--35\,677, Sep. 2025.

\bibitem{soret2024semanticgoalorientededgecomputing}
\BIBentryALTinterwordspacing
B.~Soret, I.~Leyva-Mayorga, A.~M. Mercado-Martínez, M.~Moretti,
  A.~Jurado-Navas, M.~Martinez-Gost, C.~S. de~Miguel, A.~Salas-Prendes, and
  P.~Popovski, ``Semantic and goal-oriented edge computing for satellite earth
  observation,'' 2024. [Online]. Available:
  \url{https://arxiv.org/abs/2408.15639}
\BIBentrySTDinterwordspacing

\bibitem{10644029}
H.~Zhou, Y.~Deng, X.~Liu, N.~Pappas, and A.~Nallanathan, ``Goal-oriented
  semantic communications for {6G} networks,'' \emph{IEEE Internet Things
  Mag.}, vol.~7, no.~5, pp. 104--110, Sep. 2024.

\bibitem{10016643}
J.~Chen, N.~Skatchkovsky, and O.~Simeone, ``Neuromorphic wireless cognition:
  Event-driven semantic communications for remote inference,'' \emph{IEEE
  Trans. Cogn. Commun.}, vol.~9, no.~2, pp. 252--265, Apr. 2023.

\bibitem{10388062}
Z.~Lyu, G.~Zhu, J.~Xu, B.~Ai, and S.~Cui, ``Semantic communications for image
  recovery and classification via deep joint source and channel coding,''
  \emph{IEEE Trans. Wireless Commun.}, vol.~23, no.~8, pp. 8388--8404, Aug.
  2024.

\bibitem{ren2023survey}
W.-Q. Ren, Y.-B. Qu, C.~Dong, Y.-Q. Jing, H.~Sun, Q.-H. Wu, and S.~Guo, ``A
  survey on collaborative {DNN} inference for edge intelligence,'' \emph{Mach.
  Intell. Res.}, vol.~20, no.~3, pp. 370--395, May 2023.

\bibitem{li2025taskorientedcomputationoffloadingedge}
\BIBentryALTinterwordspacing
X.~Li, S.~Bi, and Y.-J.~A. Zhang, ``Task-oriented computation offloading for
  edge inference: An integrated bayesian optimization and deep reinforcement
  learning framework,'' 2025. [Online]. Available:
  \url{https://arxiv.org/abs/2509.21090}
\BIBentrySTDinterwordspacing

\bibitem{9296560}
L.~Zeng, X.~Chen, Z.~Zhou, L.~Yang, and J.~Zhang, ``{CoEdge}: Cooperative {DNN}
  inference with adaptive workload partitioning over heterogeneous edge
  devices,'' \emph{IEEE/ACM Trans. Netw.}, vol.~29, no.~2, pp. 595--608, Apr.
  2021.

\bibitem{9837474}
J.~Shao, Y.~Mao, and J.~Zhang, ``Task-oriented communication for multidevice
  cooperative edge inference,'' \emph{IEEE Trans. Wireless Commun.}, vol.~22,
  no.~1, pp. 73--87, Jan. 2023.

\bibitem{10829586}
S.~F. Yilmaz, B.~Hasircioğlu, L.~Qiao, and D.~Gündüz, ``Private
  collaborative edge inference via over-the-air computation,'' \emph{IEEE
  Trans. Mach. Learn. Commun. Netw.}, vol.~3, pp. 215--231, Jan. 2025.

\bibitem{11121577}
Z.~Wang, A.~E. Kalør, Y.~Zhou, P.~Popovski, and K.~Huang, ``Ultra-low-latency
  edge inference for distributed sensing,'' \emph{IEEE Trans. Wireless
  Commun.}, vol.~25, pp. 1908--1922, Aug. 2026.

\bibitem{sreenivas2024llmpruningdistillationpractice}
\BIBentryALTinterwordspacing
{S. Sreenivas, et~al.}, ``{LLM} pruning and distillation in practice: The
  minitron approach,'' 2024. [Online]. Available:
  \url{https://arxiv.org/abs/2408.11796}
\BIBentrySTDinterwordspacing

\bibitem{gu2024minillm}
Y.~Gu, L.~Dong, F.~Wei, and M.~Huang, ``Mini{LLM}: Knowledge distillation of
  large language models,'' in \emph{Proc. Int. Conf. Learning Representations
  (ICLR)}, 2024.

\bibitem{10.5555/3045390.3045690}
D.~D. Lin, S.~S. Talathi, and V.~S. Annapureddy, ``Fixed point quantization of
  deep convolutional networks,'' in \emph{Proc. Int. Conf. Machine Learning},
  2016, pp. 2849--2858.

\bibitem{zhou2017incremental}
A.~Zhou, A.~Yao, Y.~Guo, L.~Xu, and Y.~Chen, ``Incremental network
  quantization: Towards lossless {CNNs} with low-precision weights,'' in
  \emph{Proc. ICLR}, 2017.

\bibitem{7780459}
K.~He, X.~Zhang, S.~Ren, and J.~Sun, ``Deep residual learning for image
  recognition,'' in \emph{Proc. IEEE/CVF Conference on Computer Vision and
  Pattern Recognition (CVPR)}, 2016, pp. 770--778.

\bibitem{10.5555/3600270.3601002}
Z.~Tong, Y.~Song, J.~Wang, and L.~Wang, ``{VideoMAE}: masked autoencoders are
  data-efficient learners for self-supervised video pre-training,'' in
  \emph{Proc. Conference on Neural Information Processing Systems (NIPS)},
  2022.

\bibitem{devlin-etal-2019-bert}
J.~Devlin, M.-W. Chang, K.~Lee, and K.~Toutanova, ``{BERT}: Pre-training of
  deep bidirectional transformers for language understanding,'' in \emph{Proc.
  Conference of the North {A}merican Chapter of the Association for
  Computational Linguistics: Human Language Technologies}, Jun. 2019, pp.
  4171--4186.

\bibitem{10.5555/3618408.3619222}
J.~Li, D.~Li, S.~Savarese, and S.~Hoi, ``{BLIP-2}: Bootstrapping language-image
  pre-training with frozen image encoders and large language models,'' in
  \emph{Proc. ICML}, 2023.

\bibitem{wang2022git}
J.~Wang, Z.~Yang, X.~Hu, L.~Li, K.~Lin, Z.~Gan, Z.~Liu, C.~Liu, and L.~Wang,
  ``{GIT}: A generative image-to-text transformer for vision and language,''
  \emph{Trans. Mach. Learn. Res.}, Dec. 2022.

\bibitem{10.5555/3495724.3495883}
{T. Brown, et~al.}, ``Language models are few-shot learners,'' in \emph{Proc.
  NIPS}, 2020.

\bibitem{cover1999elements}
T.~M. Cover, \emph{Elements of Information Theory}.\hskip 1em plus 0.5em minus
  0.4em\relax John Wiley \& Sons, 1999.

\bibitem{733495}
A.~Ortega and K.~Ramchandran, ``Rate-distortion methods for image and video
  compression,'' \emph{IEEE Signal Process. Mag.}, vol.~15, no.~6, pp. 23--50,
  Nov. 1998.

\bibitem{Lin2014COCO}
T.~Lin, M.~Maire, S.~Belongie, L.~Bourdev, R.~Girshick, J.~Hays, P.~Perona,
  D.~Ramanan, C.~L. Zitnick, and P.~Doll{\'a}r, ``Microsoft {COCO}: Common
  objects in context,'' in \emph{Proc. European Conference on Computer Vision},
  2014.

\bibitem{Wang2019VaTeX}
X.~Wang, J.~Wu, J.~Chen, L.~Li, Y.~Wang, and W.~Y. Wang, ``{VaTeX}: A
  large-scale, high-quality multilingual dataset for video-and-language
  research,'' in \emph{Proc. Int. Conf. Computer Vision}, 2019.

\end{thebibliography}

\end{document}